\documentclass[letterpaper, 10 pt, conference]{ieeeconf}  

\IEEEoverridecommandlockouts                              

\usepackage[utf8]{inputenc}

\usepackage[hidelinks]{hyperref}
\usepackage[style=ieee,hyperref,natbib=true,backend=bibtex,firstinits,doi=false,%
     mincitenames=1,maxcitenames=2,maxbibnames=99,sorting=none,terseinits=false,hyperref=true]{biblatex}
\bibliography{references.bib}
\renewbibmacro*{bbx:savehash}{}
\defbibheading{bibliography}[\bibname]{\section*{References}}

\usepackage{url}
\usepackage{graphicx}
\usepackage[usenames,dvipsnames]{xcolor}
\usepackage[draft,inline,nomargin]{fixme}
\fxsetup{theme=color}

\usepackage{amsmath}
\usepackage{amssymb}
\usepackage{textcomp}
\usepackage{siunitx}

\usepackage{booktabs}
\usepackage{threeparttable}
\usepackage{multirow}

\usepackage[capitalize]{cleveref}

\usepackage{tikz}
\usetikzlibrary{arrows}
\usetikzlibrary{positioning,calc}
\usetikzlibrary{decorations.pathreplacing}
\usetikzlibrary{decorations.markings}
\usetikzlibrary{fit}
\usetikzlibrary{shapes.callouts}
\usetikzlibrary{shapes.geometric}
\usetikzlibrary{matrix}
\usetikzlibrary{spy}
\usepackage{setspace}

\usepackage{pgfplots}
\pgfplotsset{compat=1.9}
\usepgfplotslibrary{groupplots}
\usepgfplotslibrary{units}
\usepgfplotslibrary{colorbrewer}
\usepackage{pgfplotstable}

\usepackage{balance}

\usepackage{blindtext}

\usepackage{ifthen}

\usepackage[export]{adjustbox}

\IfFileExists{graphicscache.sty}{\usepackage{graphicscache}}

\usepackage{placeins}
\usepackage{dblfloatfix}

\title{\LARGE \bf
SynPick: A Dataset for Dynamic Bin Picking Scene Understanding
}

\author{Arul Selvam Periyasamy$^{*}$, Max Schwarz$^{*}$, and Sven Behnke
\thanks{$^{*}$ Equal contribution.}
\thanks{All authors are with the Autonomous Intelligent Systems group of University of Bonn, Germany; {\tt periyasa@ais.uni-bonn.de}}%
}

\pgfplotstableread{
    Label trainpick trainmove testpick testmove trainbad testbad
    master\_chef\_can   26766  57666  7398  12258  24219  8814
    cracker\_box        28188  36324  4416  12843  38223  14730
    sugar\_box          41220  56286  13053  11586  30519  11913
    tomato\_soup\_can   40266  61023  8022  9366  47328  10695
    mustard\_bottle     44664  58788  9123  9648  40263  13989
    tuna\_fish\_can     56466  59259  8967  8292  34287  18390
    pudding\_box        57084  50361  11631  17166  44799  19437
    gelatin\_box        50478  63588  12894  13248  54705  7863
    potted\_meat\_can   38703  63066  11322  6990  50628  21957
    banana              56925  51960  12729  6258  36954  15726
    pitcher\_base       22221  70230  3156  10158  71175  10746
    bleach\_cleanser    33363  45567  8337  11121  64764  15711
    bowl                39948  54078  9282  8583  49191  20502
    mug                 39321  57417  11172  13566  55908  16125
    power\_drill        46404  61002  16980  18003  72078  12774
    wood\_block         21243  58026  6774  21495  74916  19017
    scissors            60675  72261  8331  9213  35655  12408
    large\_marker       38208  55785  12891  9924  61554  18381
    large\_clamp        57849  59655  13209  17319  49998  14148
    extra\_large\_clamp 54249  50304  12258  15348  38325  14091
    foam\_brick         33195  47892  8808  20448 86898  16185
}\testdata

\begin{document}

\maketitle

\begin{abstract}

  We present SynPick, a synthetic dataset for dynamic scene understanding
  in bin-picking scenarios.
  In contrast to existing datasets, our dataset is both situated in a realistic
  industrial application domain---inspired by the well-known Amazon Robotics
  Challenge (ARC)---and features dynamic scenes with authentic picking actions
  as chosen by our picking heuristic developed for the ARC 2017.
  The dataset is compatible with the popular BOP dataset format.
  We describe the dataset generation process in detail, including
  object arrangement generation and manipulation simulation using the NVIDIA PhysX
  physics engine.
  To cover a large action space, we perform untargeted and targeted
  picking actions, as well as random moving actions.
  To establish a baseline for object perception, a state-of-the-art pose estimation approach
  is evaluated on the dataset.
  We demonstrate the usefulness of
  tracking poses during manipulation instead of single-shot estimation
  even with a naive filtering approach.
  The generator source code and dataset are publicly available.
\end{abstract}


\section{Introduction}

6D pose estimation is an important and effective perceptual tool in many robotic
applications, such as grasping (both in industrial as well as service robotics contexts),
state estimation, and prediction. It explains scene geometry using few parameters.

Learning 6D pose estimation is tricky, however. There are issues with the
problem definition itself, such as how to deal with object symmetries and other
ambiguities, or with the chosen parametrization of the SE(3) group.
A more practical issue is how to gather training data for pose estimation.
Whereas manual annotations for classification, object detection, and even
semantic segmentation can be done in reasonable time, pose annotations are both
time-consuming and prone to annotation errors.

To address these issues, it has become common practice to augment smaller-scale real datasets
with larger synthetic datasets~\citep{hodan2018bop}. We present such a synthetic
dataset specifically focused on bin picking. Our immediate inspiration is the
setting of the Amazon Robotics Challenge 2017, which required participants
to do both \textit{targeted} picking of desired objects, and \textit{untargeted}
emptying of totes.

\begin{figure}
 \centering \setlength{\fboxsep}{0pt} \setlength{\tabcolsep}{1pt} \footnotesize
 \begin{tabular}{cc}
  \fbox{\includegraphics[width=.5\linewidth]{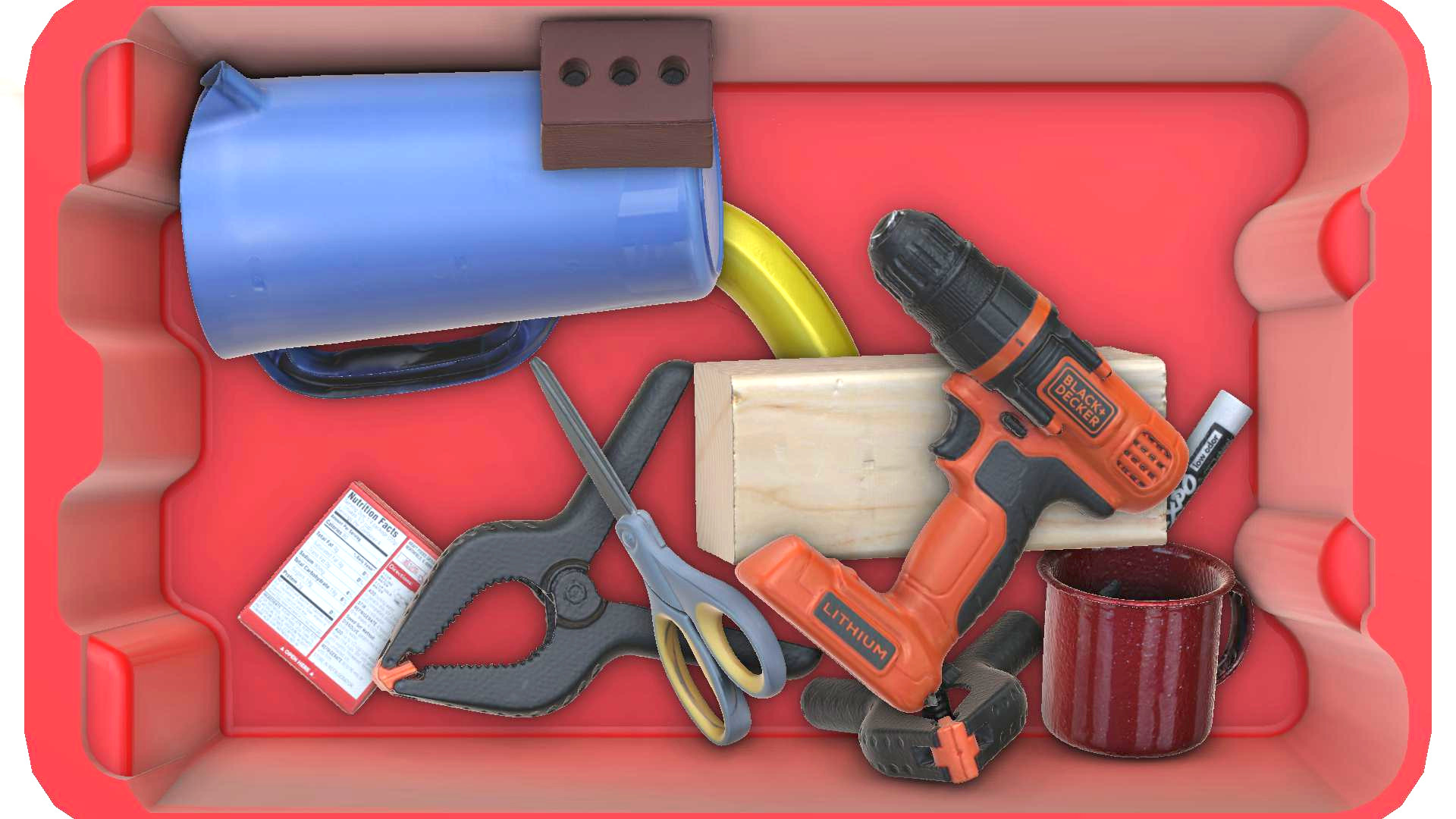}} &
  \fbox{\includegraphics[width=.5\linewidth]{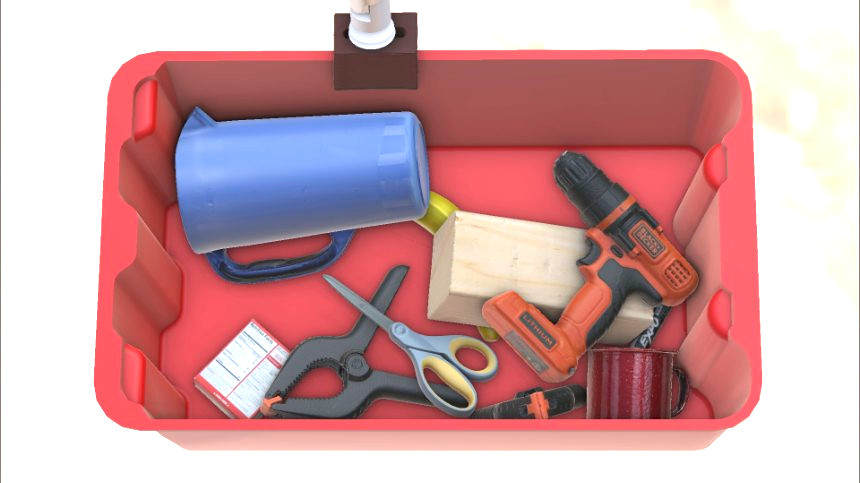}} \\
  
  (a) Initial scene & (b) Picking action \\
 
  \fbox{\includegraphics[width=.5\linewidth]{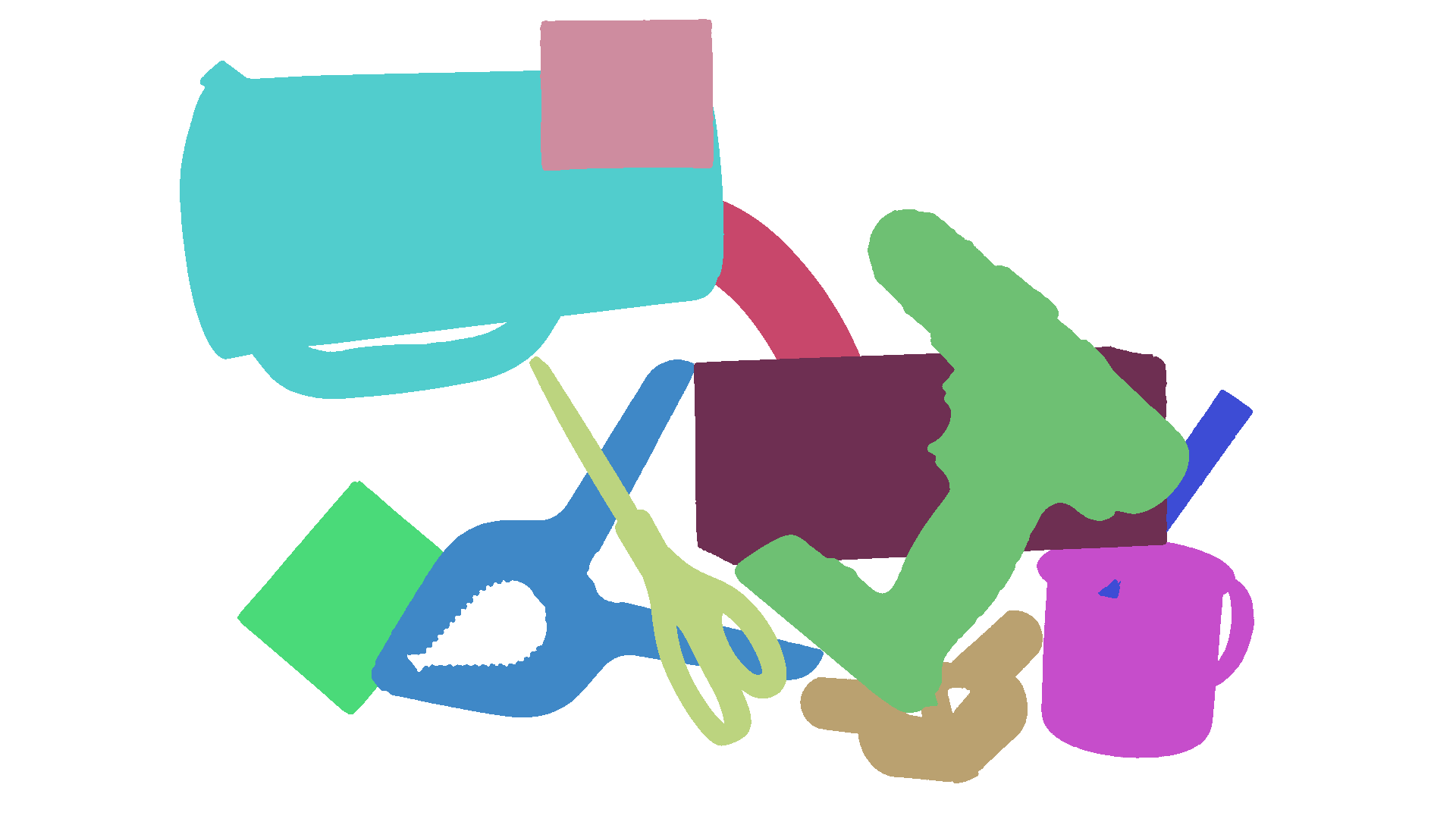}} &
  \fbox{\includegraphics[width=.5\linewidth]{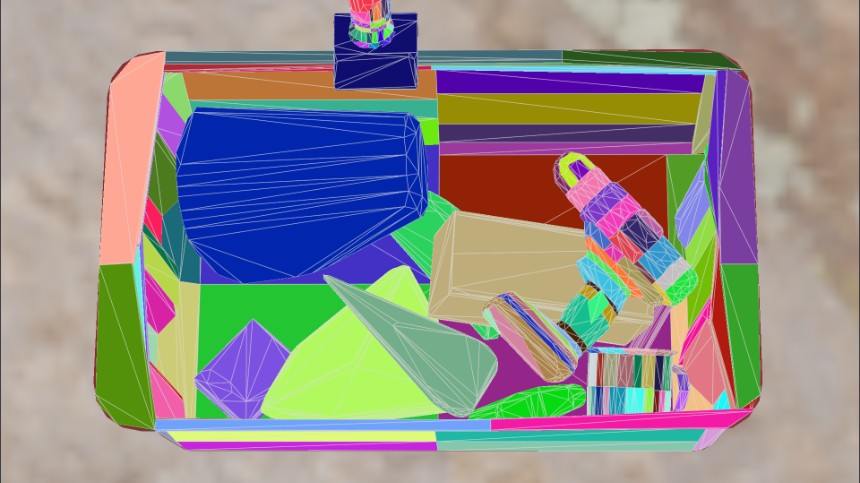}} \\
  
  (c) Semantic annotation & (d) Physics simulation \\
 \end{tabular}
 \caption{SynPick contains typical dynamic bin picking sequences with pose \& semantic segmentation annotations.}
 \label{fig:teaser}
 \vspace{-3ex}
\end{figure}

In contrast to existing datasets, our SynPick dataset does not only have static scenes, but consists of
fully dynamic picking sequences (see \cref{fig:teaser}). It is our hope that having such a dataset
(and the accompanying generator) will enable the bin picking community to
advance from analysis of static scenes to live object tracking during manipulation.\vfill\null

\noindent
In short, our contributions include:
\begin{enumerate}
 \item A dynamic scene generator, capable of producing realistic picking
   and moving sequences, both online and offline,
 \item a larger-scale multi-view dataset produced with said generator and detailed analysis
   of its properties, and
 \item a baseline evaluation of a state-of-the-art object pose estimation method and
   a filtering method for tracking on the dataset.
\end{enumerate}

Both generator source code and the SynPick dataset itself are publicly available\footnote{\url{http://ais.uni-bonn.de/datasets/synpick/}}.


\section{Related Work}

\begin{table*}
  \centering
  \caption{6D Pose Estimation and Tracking Datasets}
  \small
  \begin{tabular}{llrrcccr}
   \toprule
   Name                                           & Type             & Objects & \#Frames & Annotation      & Diverse Lighting & Dynamics & Multi-View \\
   \midrule                                                          
   YCB-Video~\citep{xiang2017posecnn}             & Real videos      & 21       & 133,827 & Semi-auto & Yes     & Static    & Moving cam\\ 
   Linemod-Occluded~\citep{brachmann2014learning} & Real videos      & 8        & 1,214   & Semi-auto & No     & Static    & Moving cam\\
   TUD-L~\citep{hodan2018bop}                     & Real videos      & 3        & 23,914  & Semi-auto & Yes      & Static    & Moving cam\\
   TYO-L~\citep{hodan2018bop}                     & Real images      & 21       & 1,670   & Manual    & Yes      & Static    & No \\
   HomebrewedDB~\citep{kaskman2019homebreweddb}   & Real videos      & 33       & 17,420  & Semi-auto & No       & Static    & Moving cam\\
   BlenderProc4BOP~\citep{hodavn2020bop}          & Synthetic images & flexible & 50,000  & Automated & Yes      & Static    & 25 views \\
   FAT~\citep{tremblay2018falling}                & Synthetic videos & 21       & 61,500  & Automated & Yes      & Falling   & Stereo \\
   ObjectSynth~\citep{hodavn2019photorealistic}   & Synthetic images & 39       & 600,000 & Automated & Yes      & Static    & ~200 views \\
   \textbf{SynPick (ours)}                        & Synthetic videos & 21       & 503232  & Automated & Yes      & Pick/Move & 3 views \\

   \bottomrule
  \end{tabular}
  \label{tab:datasets}
 \vspace{-4mm}
 \end{table*}

In recent years, with the advent of deep learning models for 6D object pose estimation,
the datasets for training and benchmarking these models have also grown in size.
The generation of large-scale datasets for 6D object pose estimation remains work-intensive, though.
Unlike obtaining ground truth annotations for 2D computer vision tasks like object detection or object classification,
annotating 6D ground truth poses is time-consuming, making the manual annotation of 6D poses on a large scale prohibitive.
Thus, most of the datasets for benchmarking 6D object pose estimation rely on semi-automated pipelines.
These pipelines involve capturing short video sequences of tabletop scenes, annotating the first frame of the sequence manually, and propagating the annotated poses to the rest of the frames by computing the camera trajectory using visual odometry techniques.
During the training of deep learning methods for 6D pose estimation, real annotated training images are often supplemented with synthetic images.
Synthetic image generation is done based on two approaches: ``render \& paste'' and physics-based rendering (PBR). The ``render \& paste'' technique involves rendering objects onto random backgrounds using standard rasterization.
This technique is simple and fast but generates physically unrealistic, poor-quality training images.
PBR is often implemented using ray tracing, which although compute-intensive and thus time-consuming,
generates high-quality training images.

Existing datasets for 6D object pose estimation include
YCB-Video~\citep{xiang2017posecnn}, Linemod-Occluded~\citep{brachmann2014learning},
HomebrewedDB~\citep{kaskman2019homebreweddb}, TUD-Light (TUD-L)~\citep{hodan2018bop},
Toyota Light (TYO-L)~\citep{hodan2018bop}, and Rutgers APC~\citep{rennie2016dataset}.
See \cref{tab:datasets} for an overview.
Each of the datasets focuses on specific real-world scenarios. \citet{hodavn2020bop} unified the existing datasets to a common BOP Dataset. Furthermore, they supplemented each dataset with PBR training images as a part of the BOP Challenge for benchmarking the progress in 6D object pose estimation research.
One of the key findings from the BOP Challenge 2020 is that 50K PBR images yield better results than 1M ``render \& paste'' images.
This finding provides a compelling motivation to develop efficient pipelines for physics-based training image generation.
Despite the advantages, the time-consuming nature of the ray tracing methods limits their applicability. \citet{schwarz2020stillleben} introduced Stillleben, an efficient data generation pipeline that can generate PBR training images using OpenGL rasterization on the fly,
completely eliminating the need for an offline data generation pipeline and demonstrated the advantages of synthetic data generation for semantic segmentation tasks.

An interesting orthogonal approach to supervised pose estimation is self-supervised 6D object pose estimation proposed by \citet{deng2020self}. In this approach, a 6D pose estimation module trained in a supervised manner is used to initialize the 6D poses of objects in the scene.
The robot changes the configuration of the objects in the scene by random pick and place actions. By capturing the images of objects in new configurations and propagating the initial pose estimate to these using forward kinematics, the authors generate new training data and refine the pose estimator actively.
While this method provides a scalable approach to train pose estimators, it is limited by the simple object manipulation actions the robot can perform without breaking the object pose tracking module. Our proposed method does not suffer from this limitation and can model complex object interactions accurately.
Furthermore, our data generator runs faster than real-time and can be easily parallelized.

Nearly all datasets geared towards 6D pose estimation feature only static scenes. 
While the camera often travels
around the arrangement, object configurations remain fixed. In contrast, our dataset features
dynamic scenes, suitable for training and evaluating not only pose estimation, but also pose tracking methods.

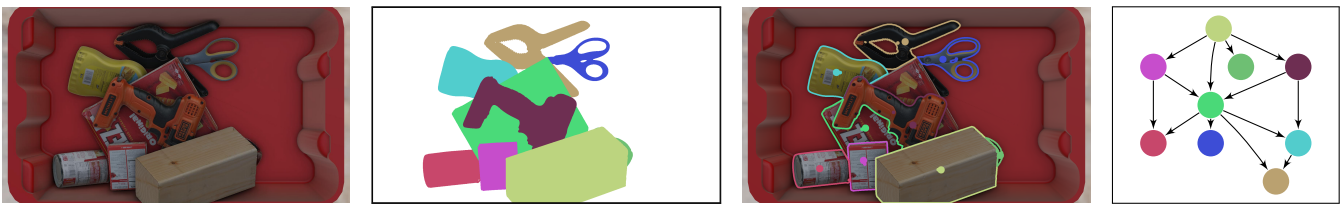
\begin{figure*}[b]
 \centering \setlength{\fboxsep}{0pt}
 \includegraphics[height=2.6cm]{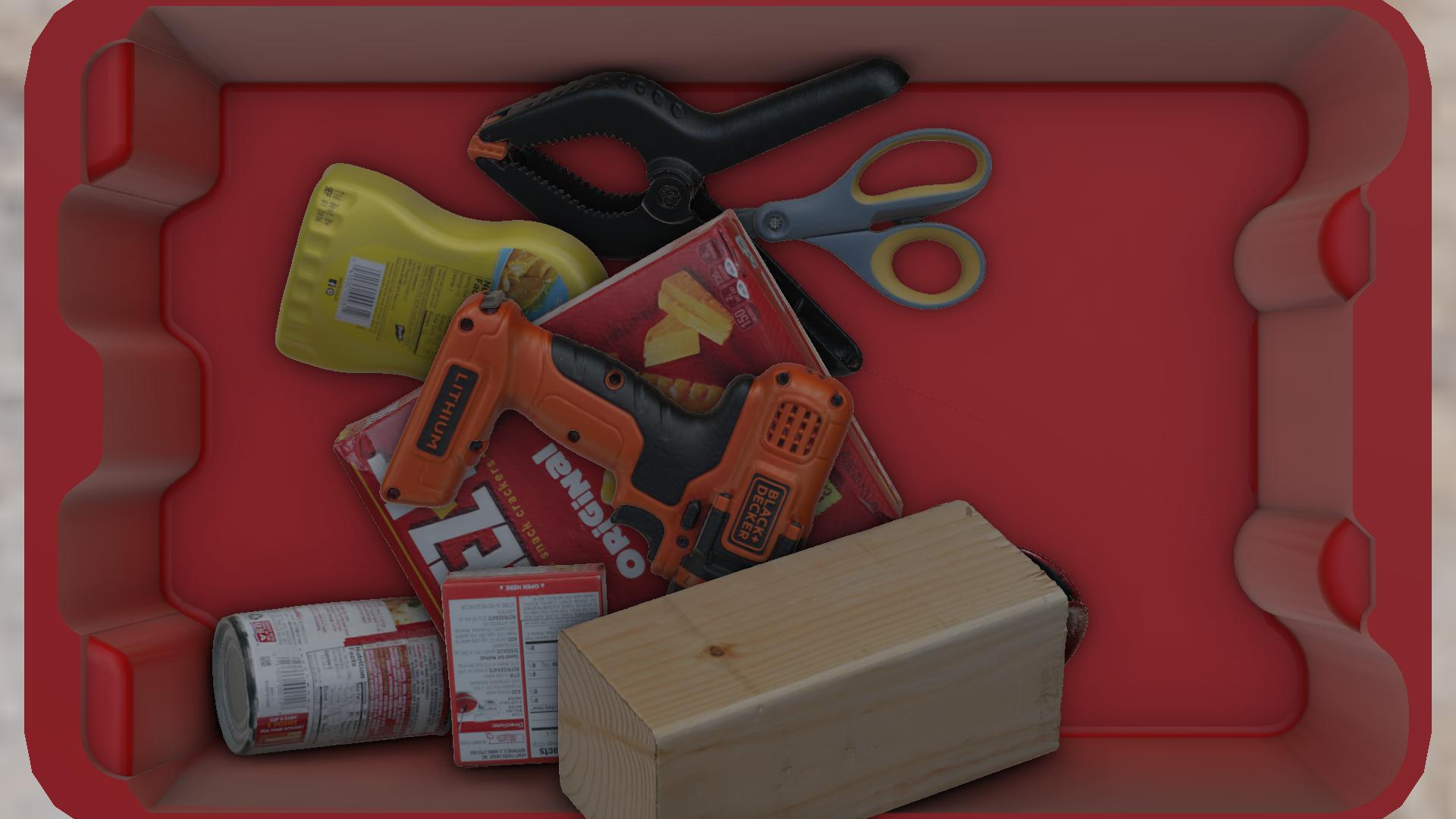}\hfill%
 \fbox{\includegraphics[frame,height=2.6cm]{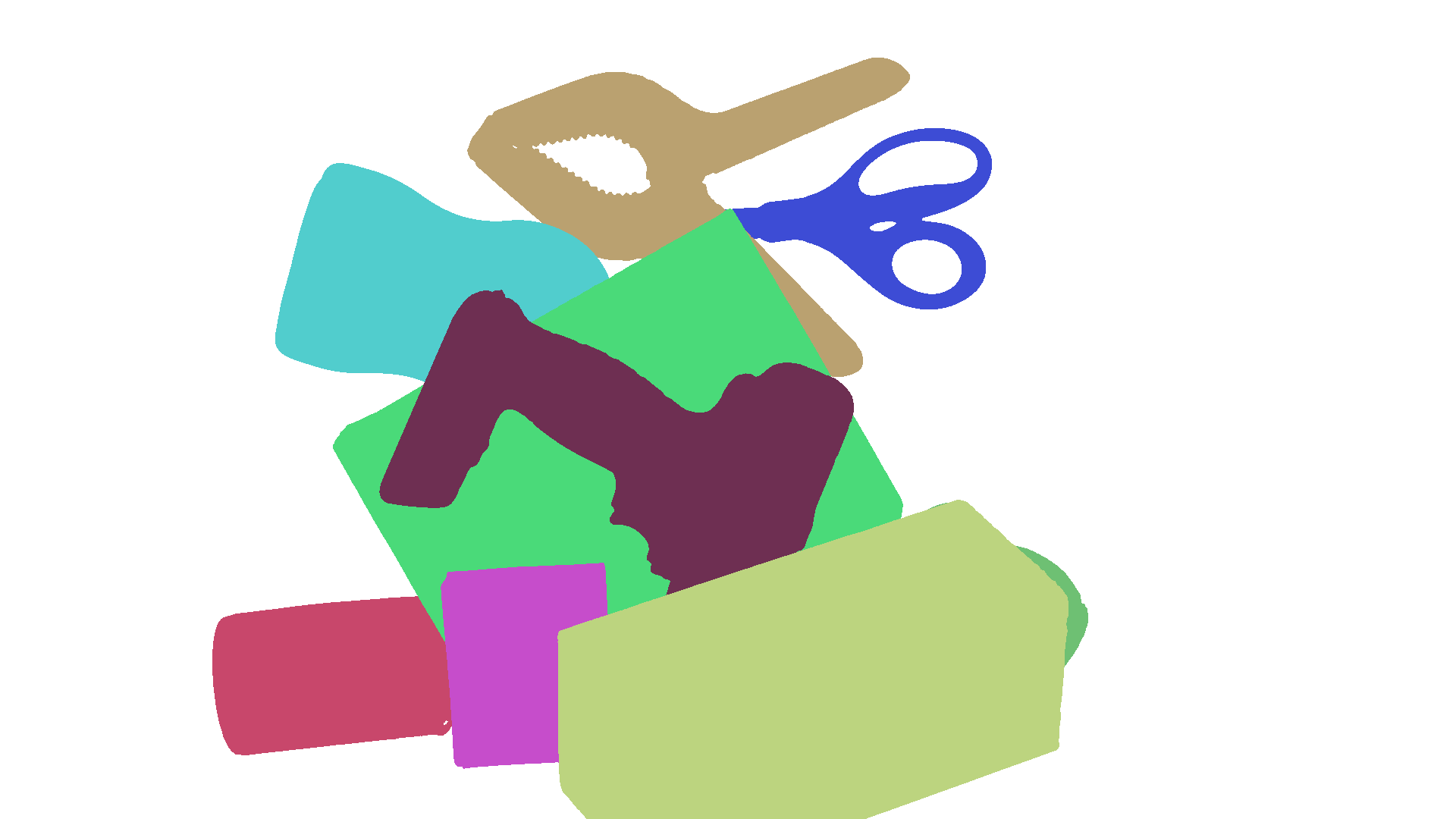}}\hfill%
 \includegraphics[height=2.6cm]{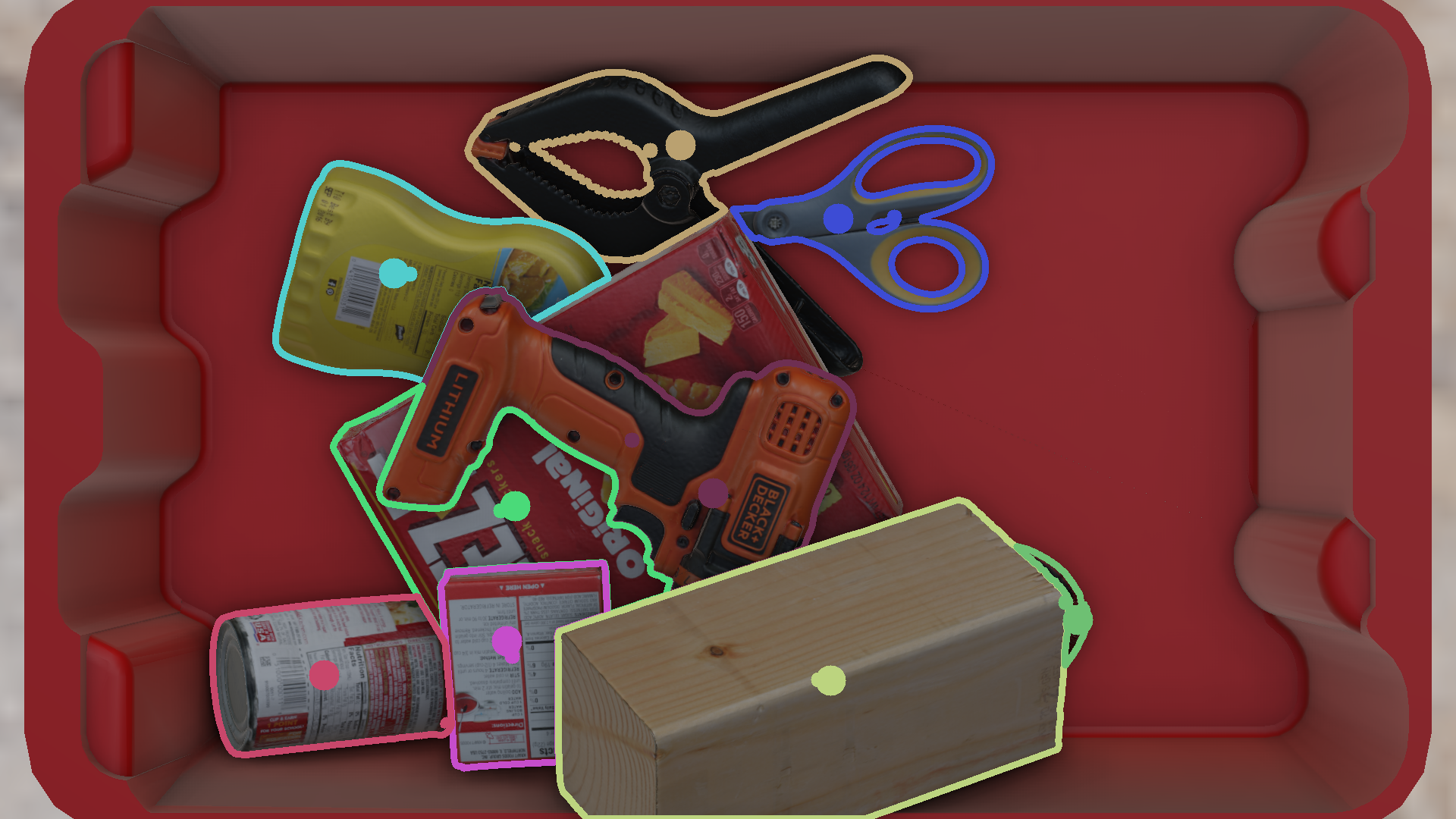}\hfill%
 \fbox{\adjustbox{minipage=[b][2.6cm][c]{3cm}}{
 \begin{center}
 
 \begin{tikzpicture}[>=latex',line join=bevel,xscale=0.3,yscale=0.2,font=\footnotesize,
 every node/.style={inner sep=0pt,text opacity=0,minimum width=10pt,minimum height=10pt}
 ]
 \definecolor{strokecolor}{rgb}{0.29,0.85,0.47};
  \node (v003_cracker_box) at (99.0bp,162.0bp) [fill=strokecolor,ellipse] {0};
  \definecolor{strokecolor}{rgb}{0.78,0.28,0.42};
  \node (v005_tomato_soup_can) at (27.0bp,90.0bp) [fill=strokecolor,ellipse] {1};
  \definecolor{strokecolor}{rgb}{0.32,0.8,0.8};
  \node (v006_mustard_bottle) at (209.0bp,90.0bp) [fill=strokecolor,ellipse] {2};
  \definecolor{strokecolor}{rgb}{0.24,0.3,0.84};
  \node (v037_scissors) at (99.0bp,90.0bp) [fill=strokecolor,ellipse] {7};
  \definecolor{strokecolor}{rgb}{0.73,0.63,0.44};
  \node (v052_extra_large_clamp) at (181.0bp,18.0bp) [fill=strokecolor,ellipse] {8};
  \definecolor{strokecolor}{rgb}{0.78,0.3,0.8};
  \node (v009_gelatin_box) at (27.0bp,234.0bp) [fill=strokecolor,ellipse] {3};
  \definecolor{strokecolor}{rgb}{0.43,0.75,0.45};
  \node (v025_mug) at (137.0bp,234.0bp) [fill=strokecolor,ellipse] {4};
  \definecolor{strokecolor}{rgb}{0.43,0.18,0.32};
  \node (v035_power_drill) at (209.0bp,234.0bp) [fill=strokecolor,ellipse] {5};
  \definecolor{strokecolor}{rgb}{0.74,0.83,0.5};
  \node (v036_wood_block) at (109.0bp,306.0bp) [fill=strokecolor,ellipse] {6};
  \draw [->] (v003_cracker_box) ..controls (74.25bp,136.94bp) and (60.476bp,123.55bp)  .. (v005_tomato_soup_can);
  \draw [->] (v003_cracker_box) ..controls (135.67bp,137.67bp) and (161.76bp,121.06bp)  .. (v006_mustard_bottle);
  \draw [->] (v003_cracker_box) ..controls (99.0bp,135.98bp) and (99.0bp,126.71bp)  .. (v037_scissors);
  \draw [->] (v003_cracker_box) ..controls (117.59bp,135.15bp) and (127.26bp,121.0bp)  .. (135.0bp,108.0bp) .. controls (147.38bp,87.192bp) and (159.9bp,62.694bp)  .. (v052_extra_large_clamp);
  \draw [->] (v006_mustard_bottle) ..controls (199.09bp,64.216bp) and (195.06bp,54.14bp)  .. (v052_extra_large_clamp);
  \draw [->] (v009_gelatin_box) ..controls (51.75bp,208.94bp) and (65.524bp,195.55bp)  .. (v003_cracker_box);
  \draw [->] (v009_gelatin_box) ..controls (27.0bp,191.67bp) and (27.0bp,147.21bp)  .. (v005_tomato_soup_can);
  \draw [->] (v035_power_drill) ..controls (172.33bp,209.67bp) and (146.24bp,193.06bp)  .. (v003_cracker_box);
  \draw [->] (v035_power_drill) ..controls (209.0bp,191.67bp) and (209.0bp,147.21bp)  .. (v006_mustard_bottle);
  \draw [->] (v036_wood_block) ..controls (104.09bp,277.53bp) and (102.04bp,264.05bp)  .. (101.0bp,252.0bp) .. controls (99.216bp,231.34bp) and (98.753bp,207.89bp)  .. (v003_cracker_box);
  \draw [->] (v036_wood_block) ..controls (80.817bp,280.94bp) and (64.309bp,266.85bp)  .. (v009_gelatin_box);
  \draw [->] (v036_wood_block) ..controls (118.91bp,280.22bp) and (122.94bp,270.14bp)  .. (v025_mug);
  \draw [->] (v036_wood_block) ..controls (142.28bp,281.7bp) and (164.74bp,265.98bp)  .. (v035_power_drill);
 \end{tikzpicture}
 \end{center}}}
 \caption{Picking heuristic. Starting from an RGB image and the ground truth
  segmentation, our ARC 2017 pipeline~\citep{schwarz2018fast} generates object
  contours with suction grasp points, as well as a clutter graph describing the
  scene layout. Each edge in the graph points from the object on top to the
  object below.}
 \label{fig:pick_heuristic}
\end{figure*}

Most related to our work is the Falling Things Dataset (FAT)~\citep{tremblay2018falling}. It consists of renderings of randomly sampled 
subsets of YCB objects placed in different indoor and outdoor scenes. Unlike most other datasets, FAT does not
include tabletop scenes where the objects are nicely arranged without significant occlusions. Instead, the objects are dropped 
from a height onto the scenery, using a physics engine to model object interactions.
However, both the context (kitchen/temple/forest) as well as the dynamics (objects falling on the background
geometry) do not really fit a bin picking application. In comparison, our dataset features a standard picking
tote and demonstrates pick and move dynamics.

\section{Dataset Generation}

We use Stillleben~\citep{schwarz2020stillleben} as the base rendering engine for generating the SynPick dataset,
as it has built-in physics simulation for table-top arrangements and is capable of online generation.
However, Stillleben was developed for fast scene generation of different scenes. In this section, we describe the
extensions to Stillleben we developed to adapt it for generation of continuous bin picking sequences
and detail the dataset generation process.

\subsection{Scene Generation}
\label{sec:scene_generaion}
We rely on Stillleben's scene arrangement engine for physically realistic scene generation.
We start with adding the tote object to the scene. Then, we randomly sample a set of objects such that the total volume of the objects does not exceed a threshold of 7\,l, which generates scenes that contain many occlusions.
We let the objects fall into the tote from a fixed height and simulate the effect of gravity and collision with other objects and the tote.
Stillleben internally uses the PhysX engine by NVIDIA\footnote{PhysX: \url{https://developer.nvidia.com/physx-sdk}} for simulation.
We use meshes provided by the YCB-Video dataset~\citep{xiang2017posecnn} for rendering.
Simulating collisions with these high-resolution meshes would be prohibitively slow. For most objects, we compute
convex hulls of low complexity.
For three highly concave objects (\texttt{cup}, \texttt{bowl}, and \texttt{power drill}), we use the V-HACD~\citep{mamou2016volumetric} convex decomposition algorithm to find a number of smaller convex meshes describing the geometry accurately.

\begin{figure}
 \centering
 \includegraphics[height=2.2cm]{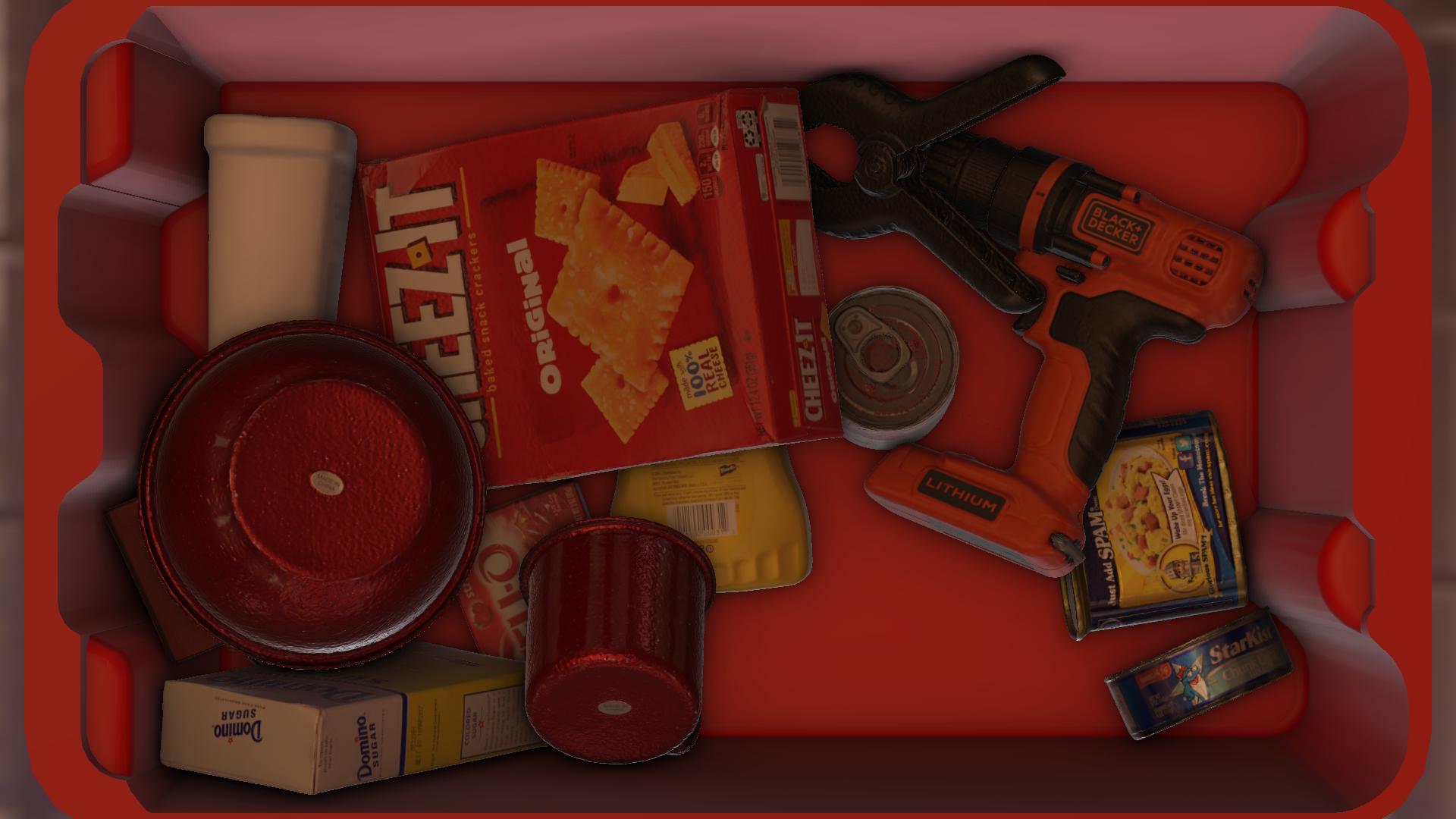}
 \includegraphics[height=2.2cm]{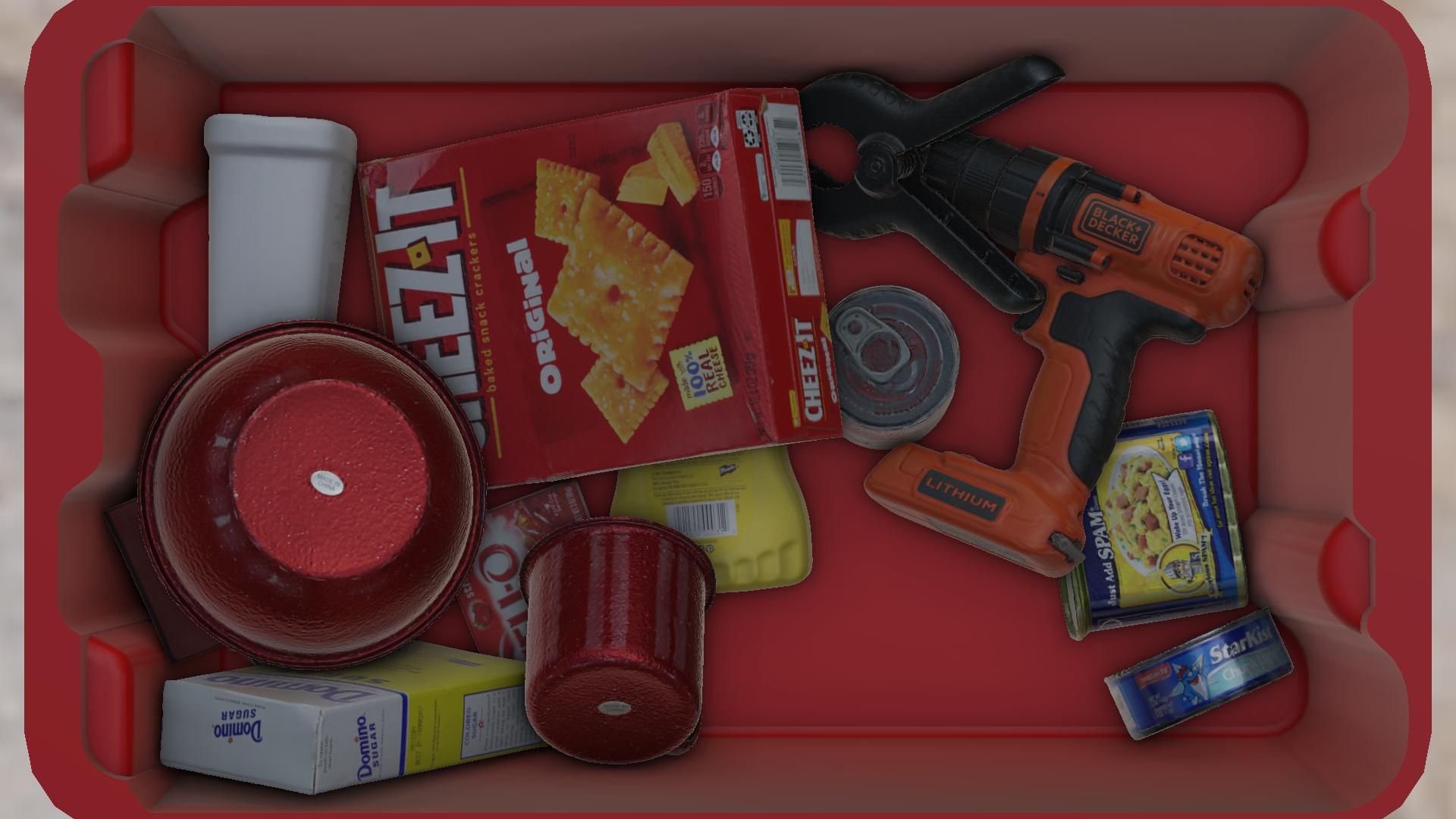}
 \includegraphics[height=2.2cm]{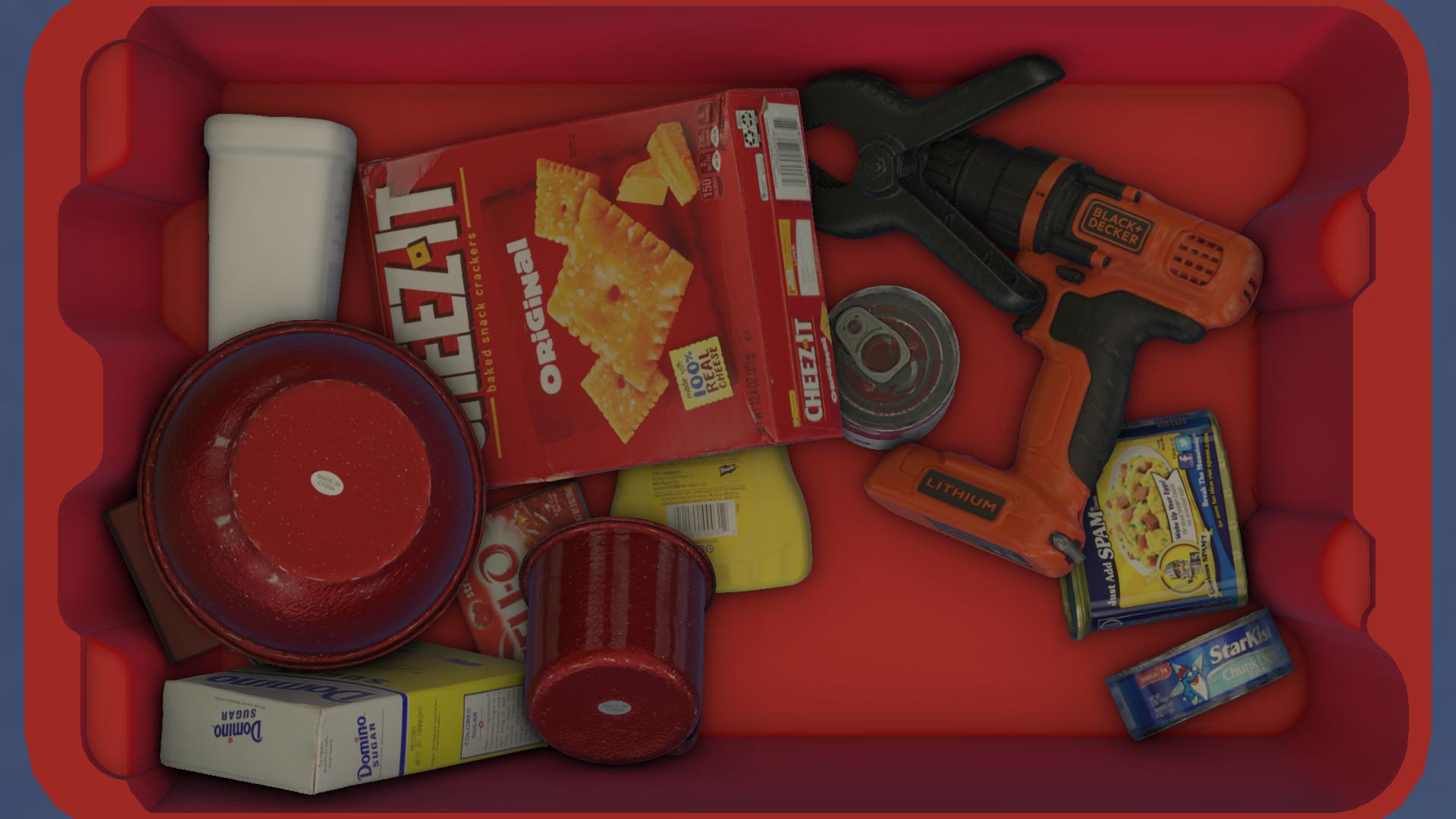}
 \includegraphics[height=2.2cm]{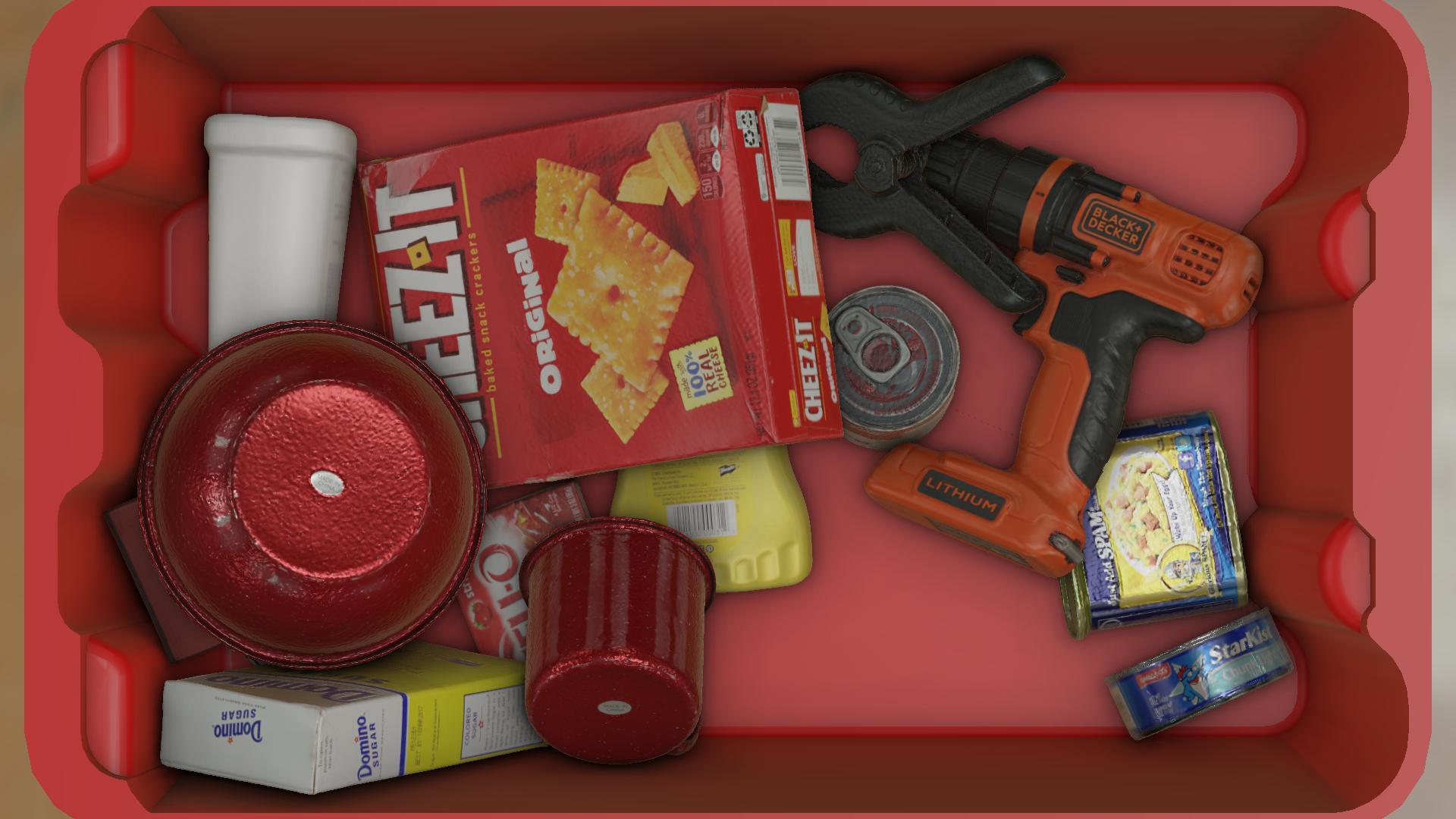}
 \includegraphics[height=2.2cm]{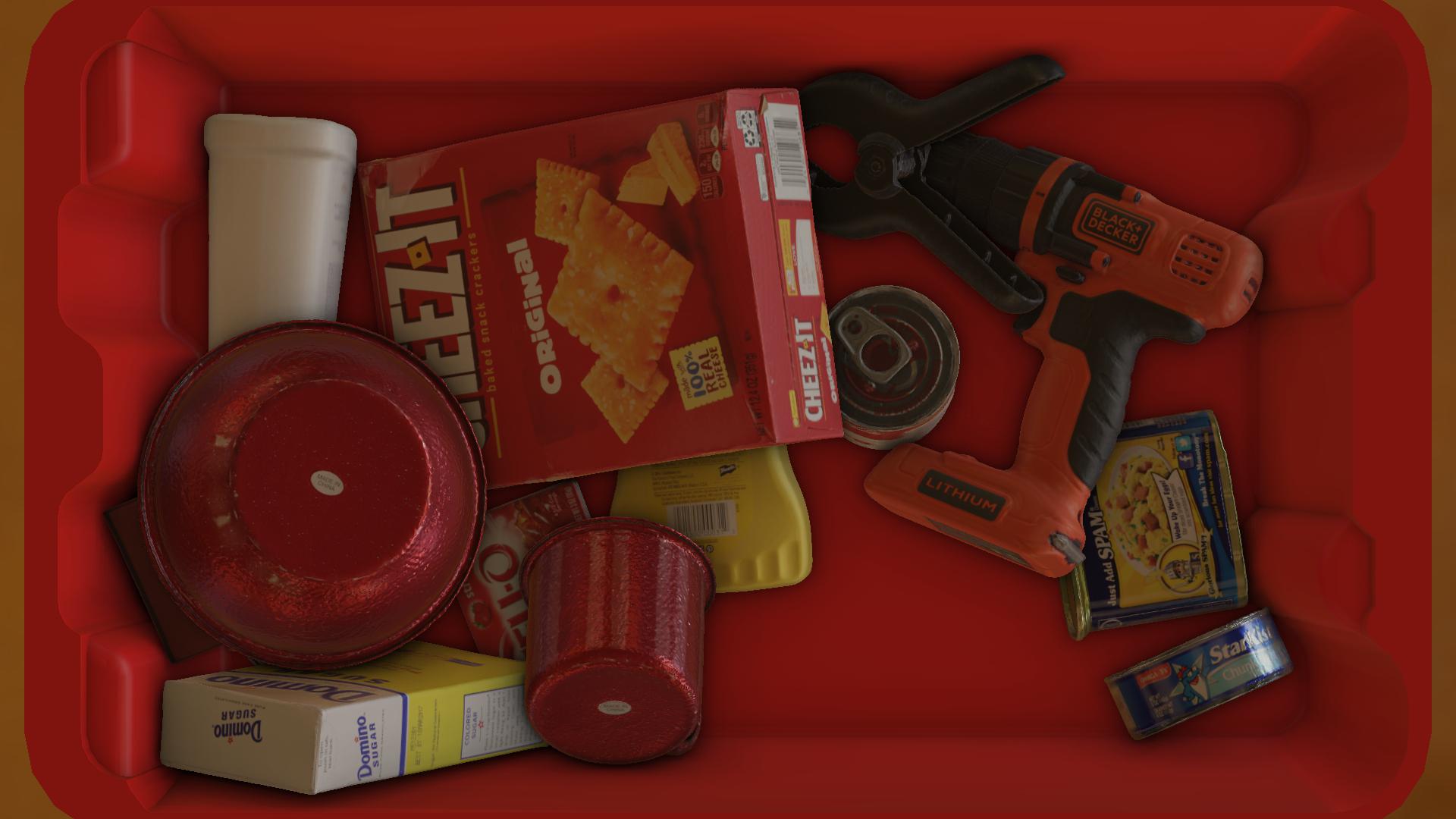}
 \includegraphics[height=2.2cm]{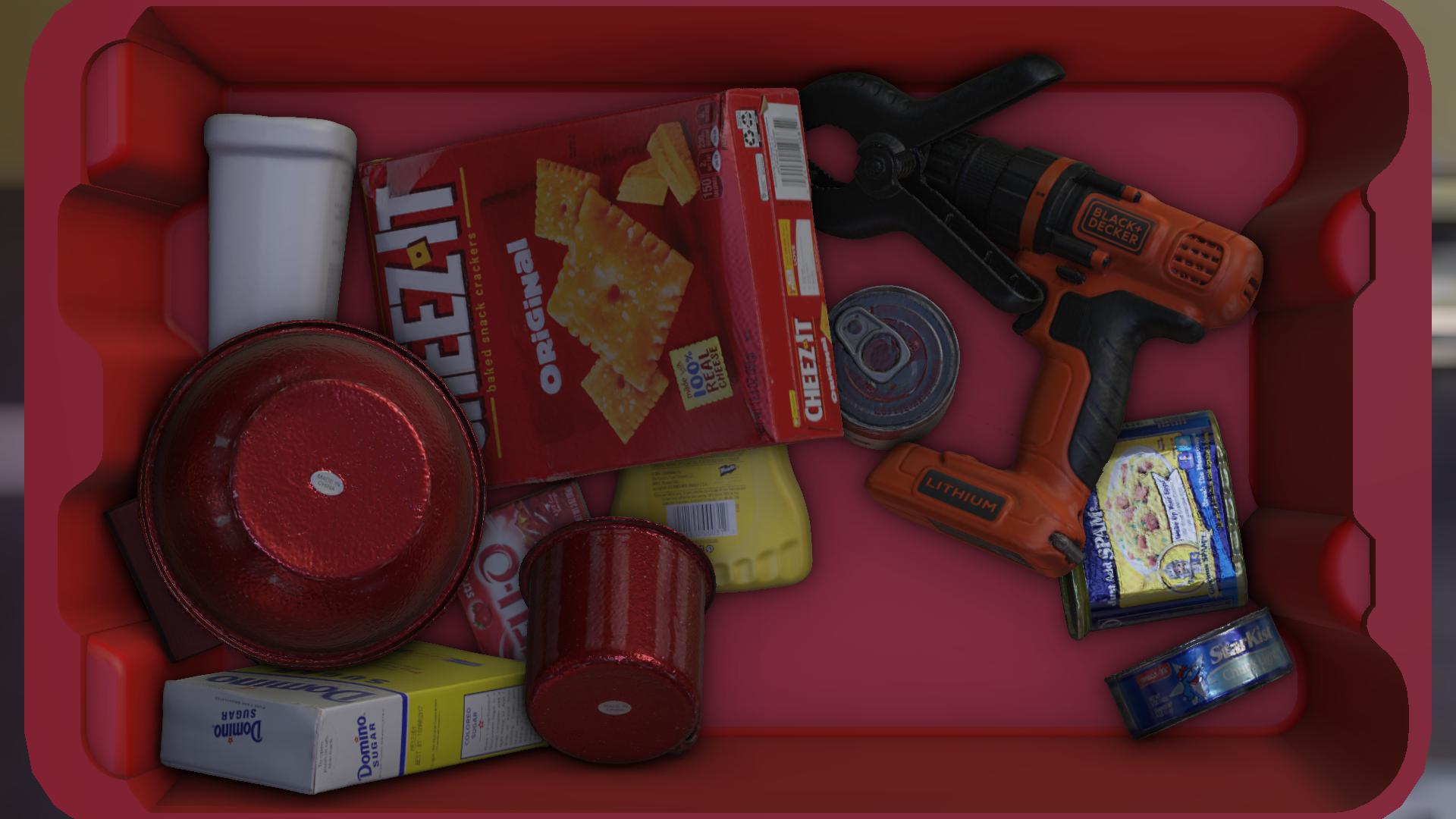}
 \caption{Lighting variants. We show the same object arrangement in different sIBL light maps.}
 \label{fig:ibl}
 \vspace{-2ex}
\end{figure}

After the scene arrangement has been generated, we start the generation of training frames. During generation,
we run PhysX simulation with a very small step size of 2\,ms to ensure realistic behavior.
With a rate of 15\,Hz (relative to simulated time), we generate output frames from three different camera poses
above the tote.
The simulated cameras have Full HD resolution (1920$\times$1080) and are mounted 2\,m above the tote, similar
to our ARC 2017 system~\citep{schwarz2018fast}.

For RGB rendering, the standard Stillleben pipeline is used. A physics-based rendering shader pipeline combined
with randomly selected image-based lighting (IBL) maps from the sIBL archive\footnote{\url{http://www.hdrlabs.com/sibl/archive.html}} ensure interesting and realistic lighting effects (see \cref{fig:ibl}).

In the following sections, we describe the simulation of pick and move actions in detail.

\subsection{Untargeted Picking}

In the pick task simulation, we simulate the robot emptying the tote by picking the objects out of the tote,
similar to the \textit{Stow} task of the Amazon Robotics Challenge 2017.
Starting with the tote filled with objects in random configurations generated by the arrangement engine,
we follow the pick planning and grasp heuristics selection strategy proposed by \citet{schwarz2018fast}. See \cref{fig:pick_heuristic} for a visualization.
We substitute the semantic segmentation network with the ground-truth segmentation masks produced by Stillleben.
After extracting object contours, ideal suction points are found inside the contour. Depending on object weight,
either the \textit{pole of inaccessibility}~\citep{garcia2007poles}, i.e. the point with maximum distance to the contour
is found to minimize the chance of catching other objects, or the center of mass is computed from the contour to ensure
good mass distribution.

The system introduced by \citet{schwarz2018fast} also computes a \textit{clutter graph}, identifying which object
is resting on top of which object. The final decision on which object to pick is based on this graph, ensuring
that we do not attempt to grasp objects caught beneath others.

\begin{figure}
   
  \centering
    \includegraphics[width=\linewidth,clip,trim=0 300 100 300]{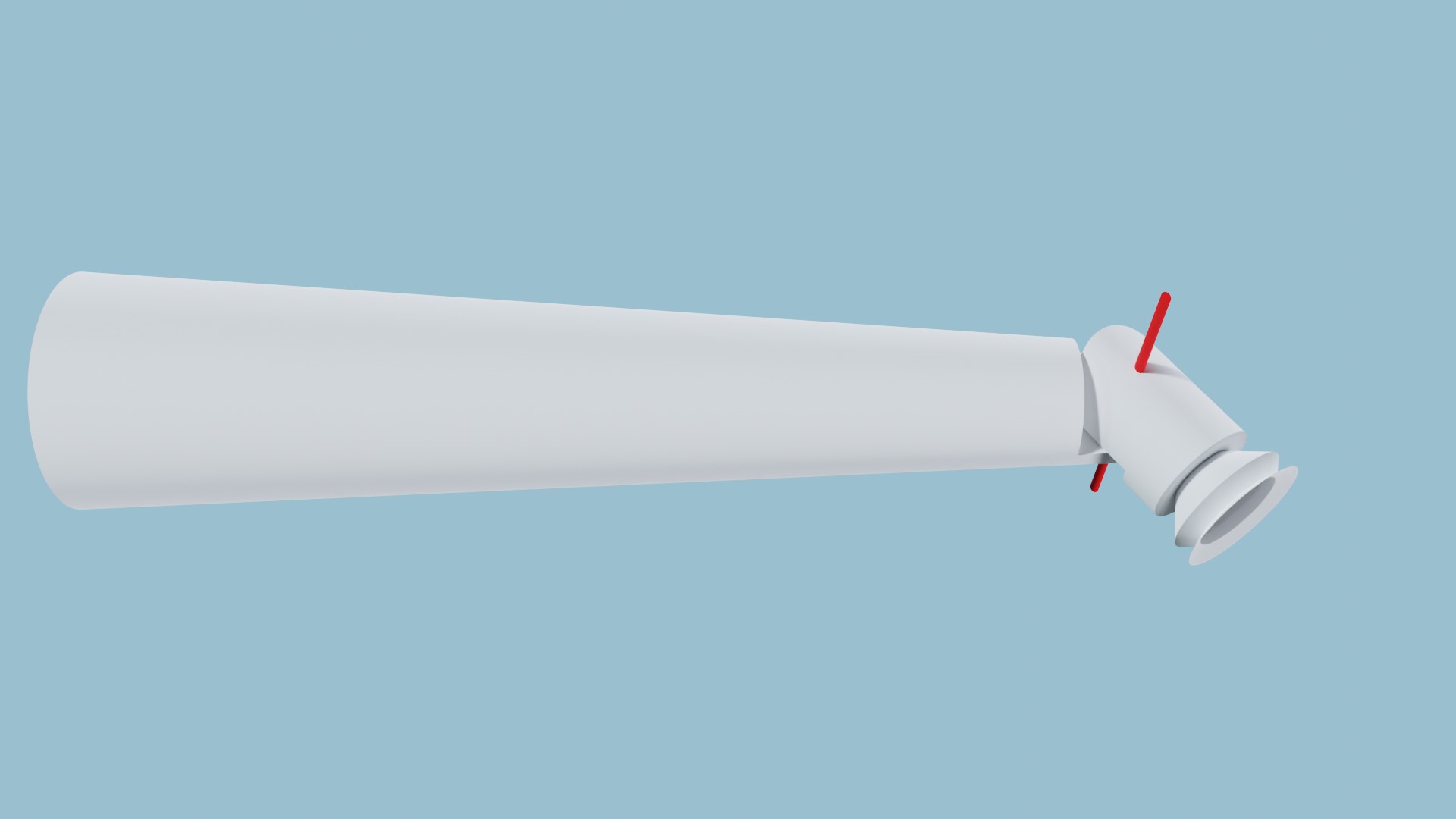}
    \caption{Articulated gripper. The cone-shaped finger
    ends in an actuated tiltable suction cup (joint axis marked in red).}
    \label{fig:gripper}
	\vspace{-4ex}
  \end{figure}

Once the target object and the corresponding suction point are determined, we find an inverse kinematics solution
for the articulated gripper (see \cref{fig:gripper}), which places the suction cup orthogonally on the local surface, as estimated
by a local average of the pixel-wise normals. A gripper trajectory is then computed to bring the gripper to
the target pose, apply suction, and lift the object out of the tote.

The gripper is moved along the trajectory using Cartesian impedance control, with a stiffness of 2500\,$\frac{N}{m}$ and
a spring damping of 200\,$\frac{Ns}{m}$. The force exerted by the impedance control is limited to 200\,$N$.
The high stiffness simulates an industrial robotic arm holding the gripper in place.

\begin{figure*}[b]
  \centering
 
 \newlength{\imgw}
 \setlength{\imgw}{3.4cm}
 \setlength{\tabcolsep}{1pt}
 \footnotesize
 \begin{tabular}{cccccc}
  \adjustbox{valign=b}{a)} &
  \includegraphics[valign=c,width=\imgw]{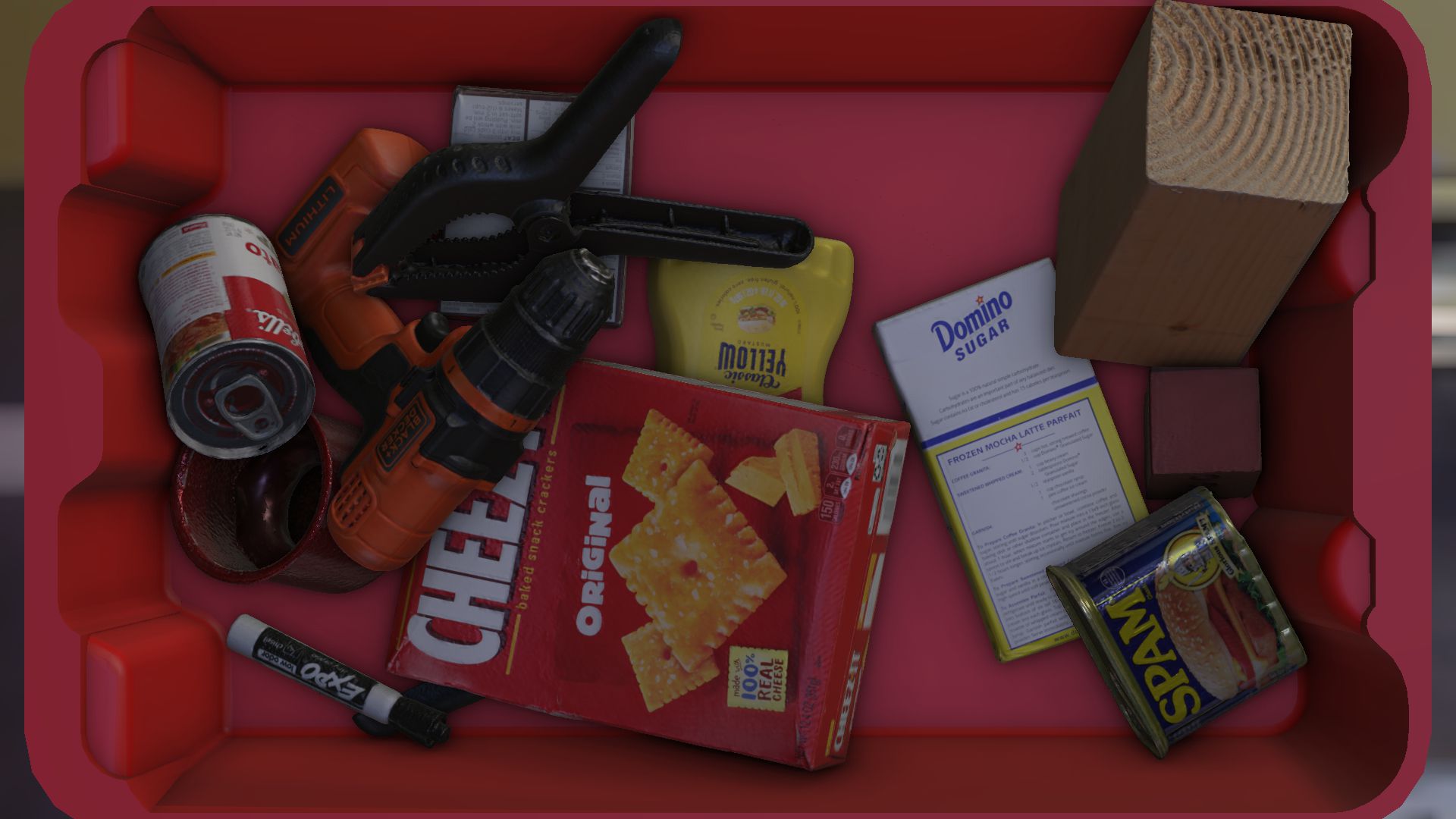} &
  \includegraphics[valign=c,width=\imgw]{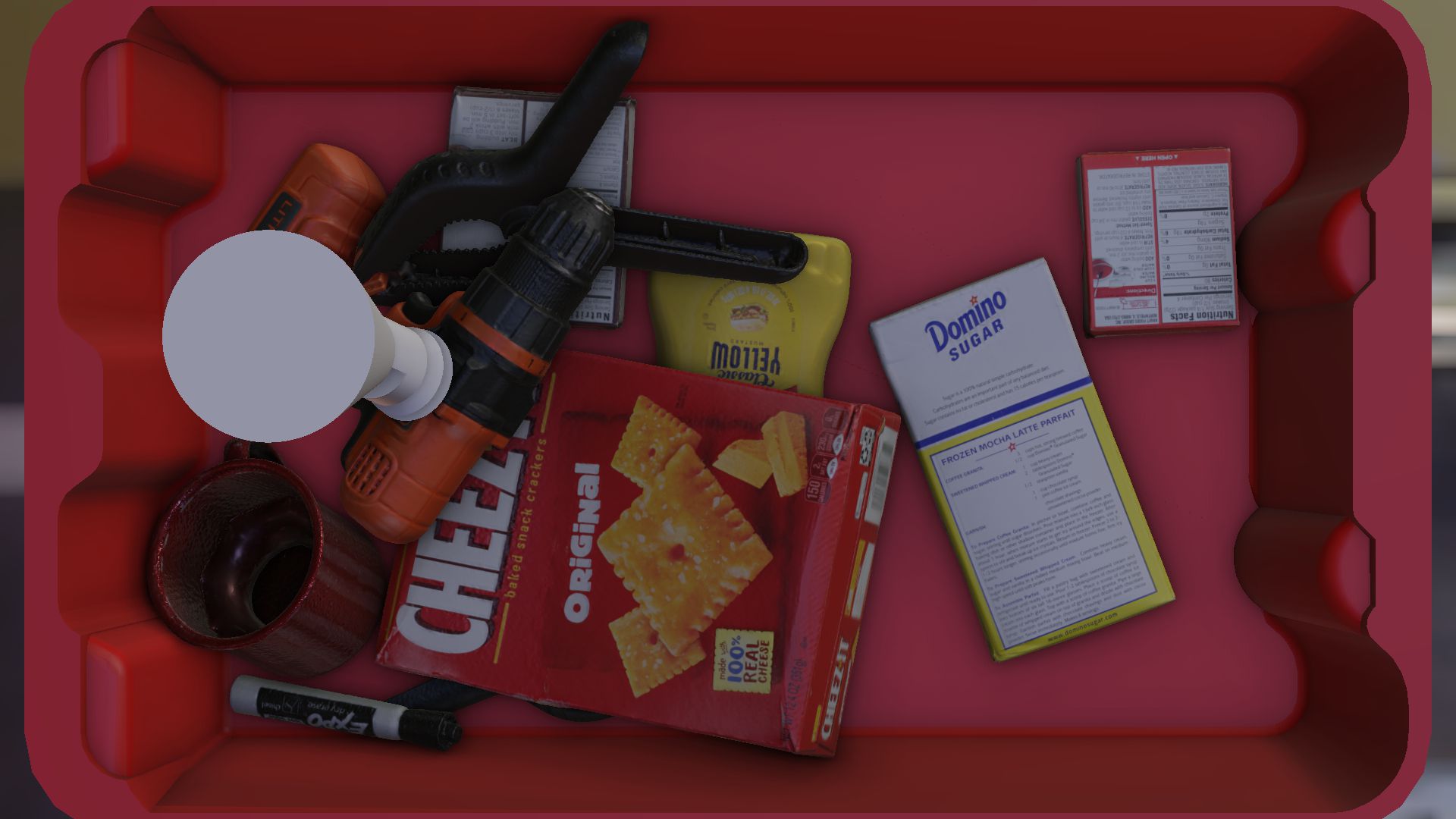} &
  \includegraphics[valign=c,width=\imgw]{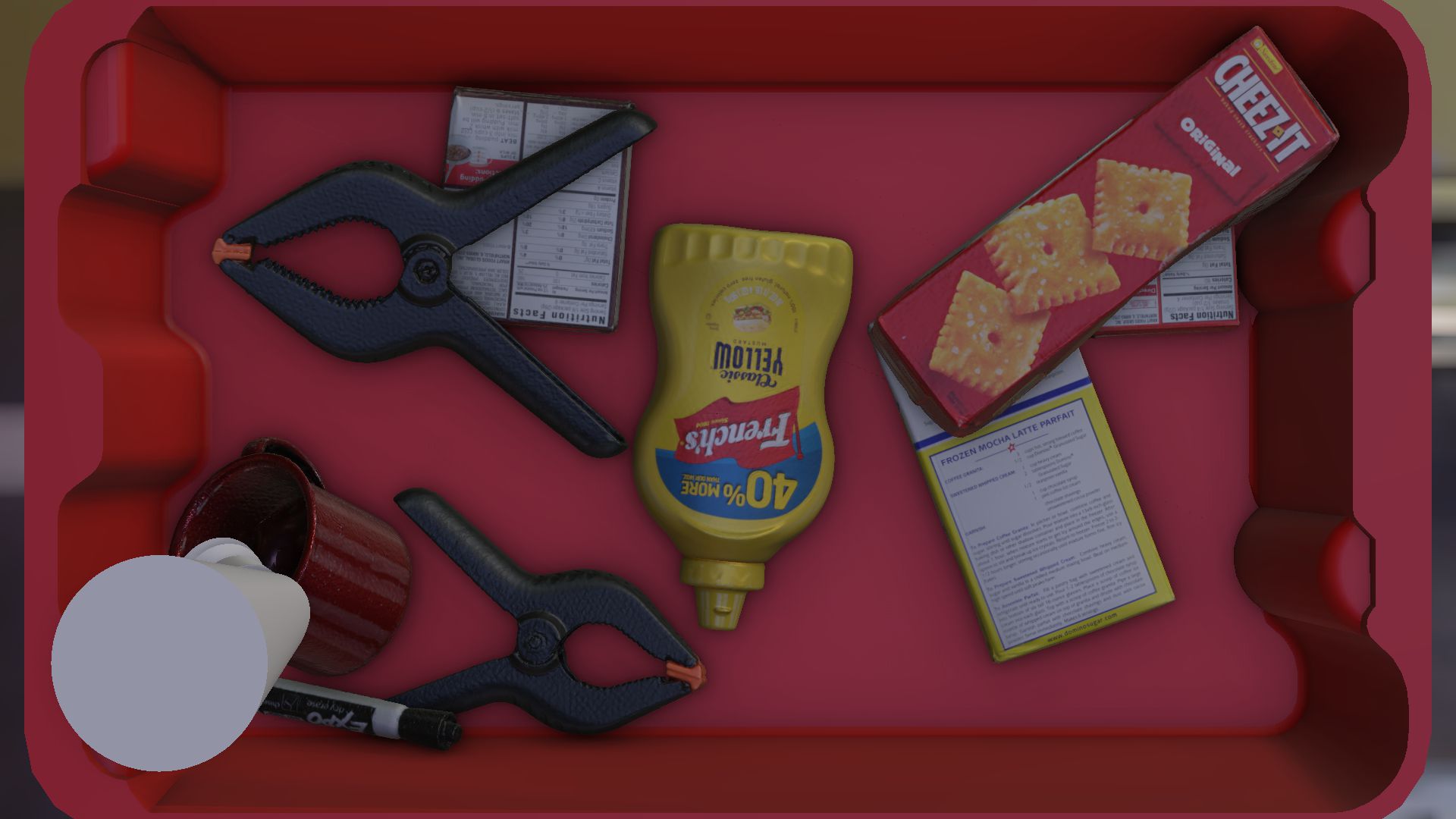} &
  \includegraphics[valign=c,width=\imgw]{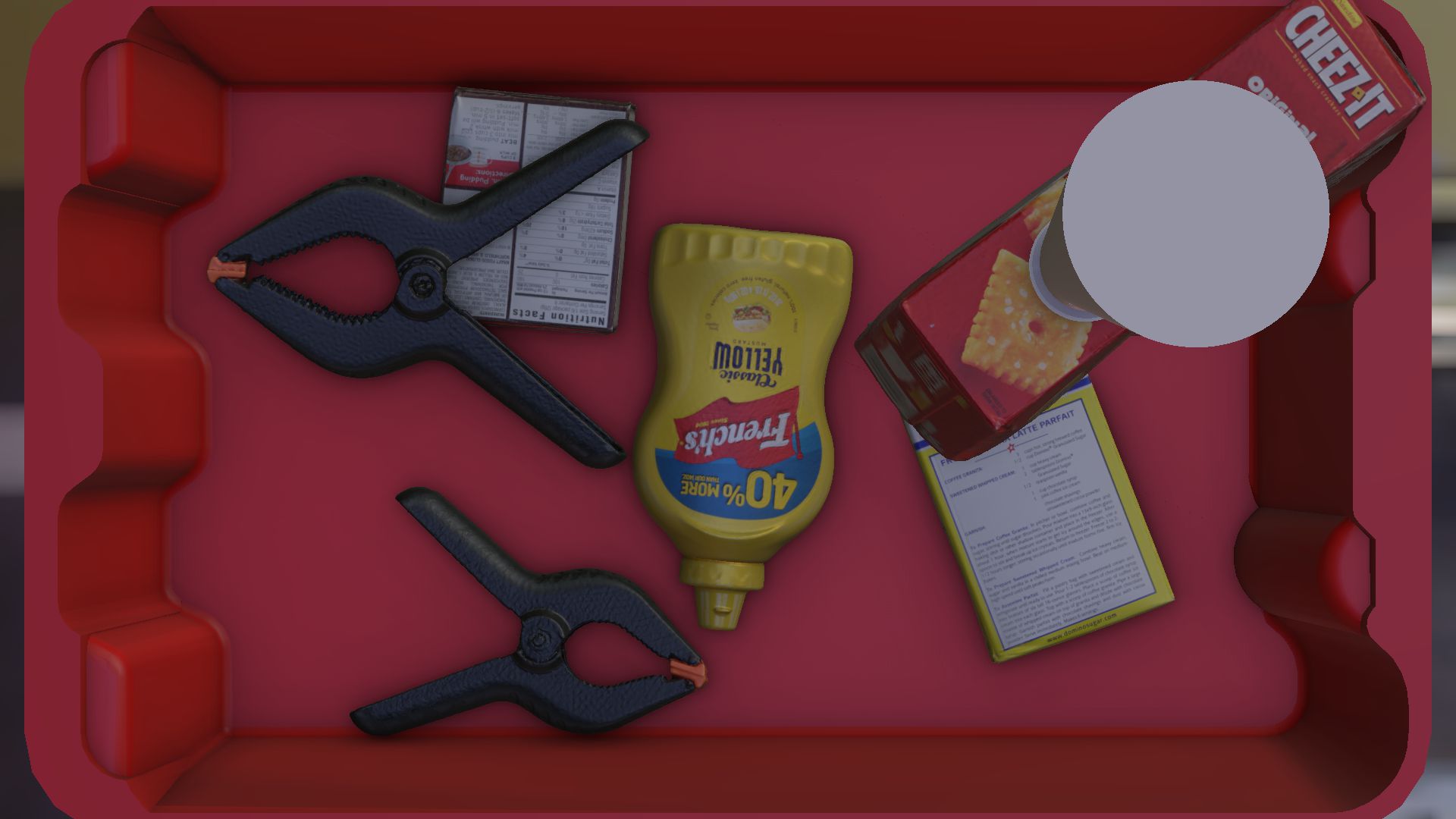} &
  \includegraphics[valign=c,width=\imgw]{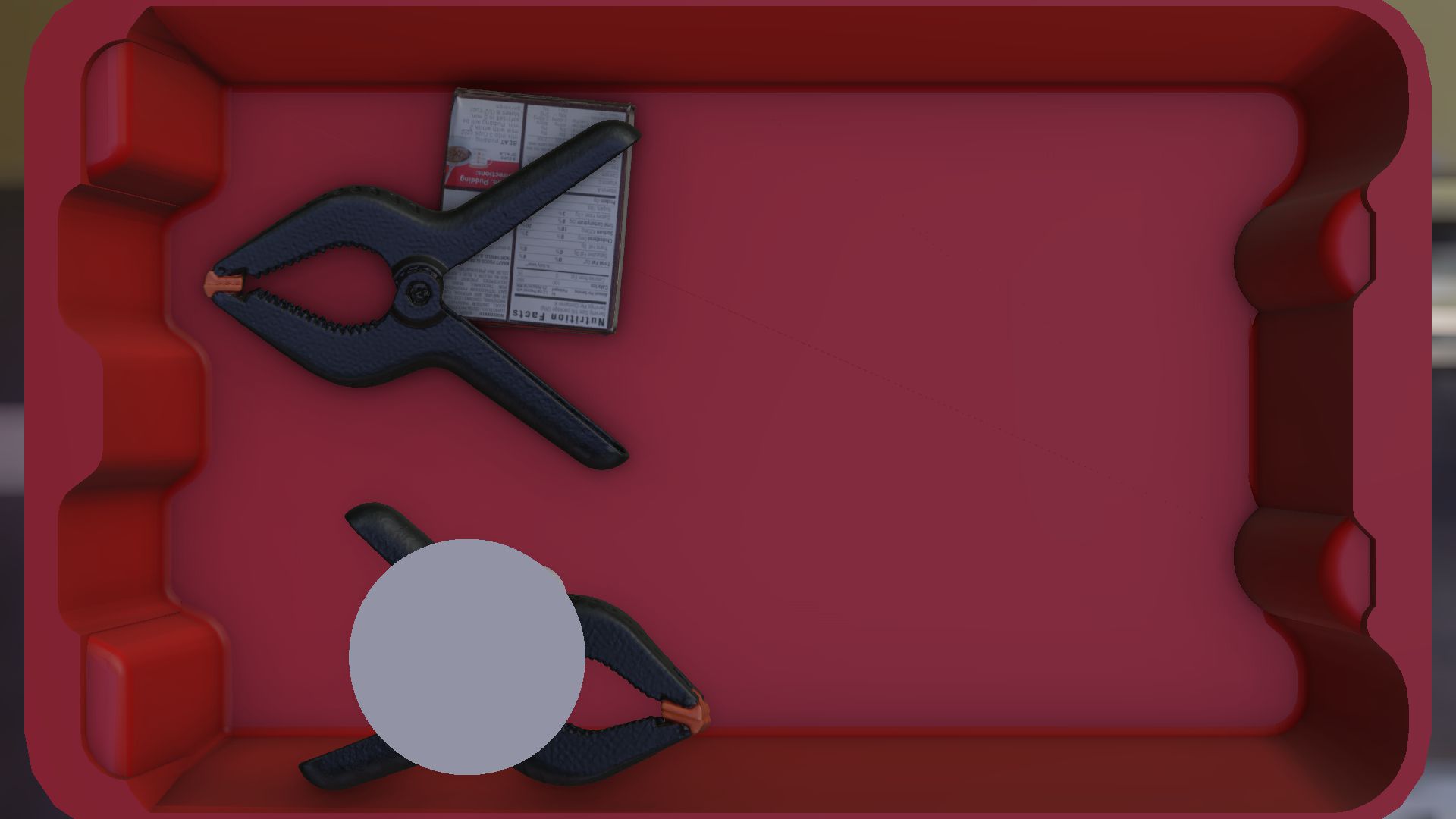} \\[0.1cm]

  \adjustbox{valign=b}{b)} &        
  \includegraphics[valign=c,width=\imgw]{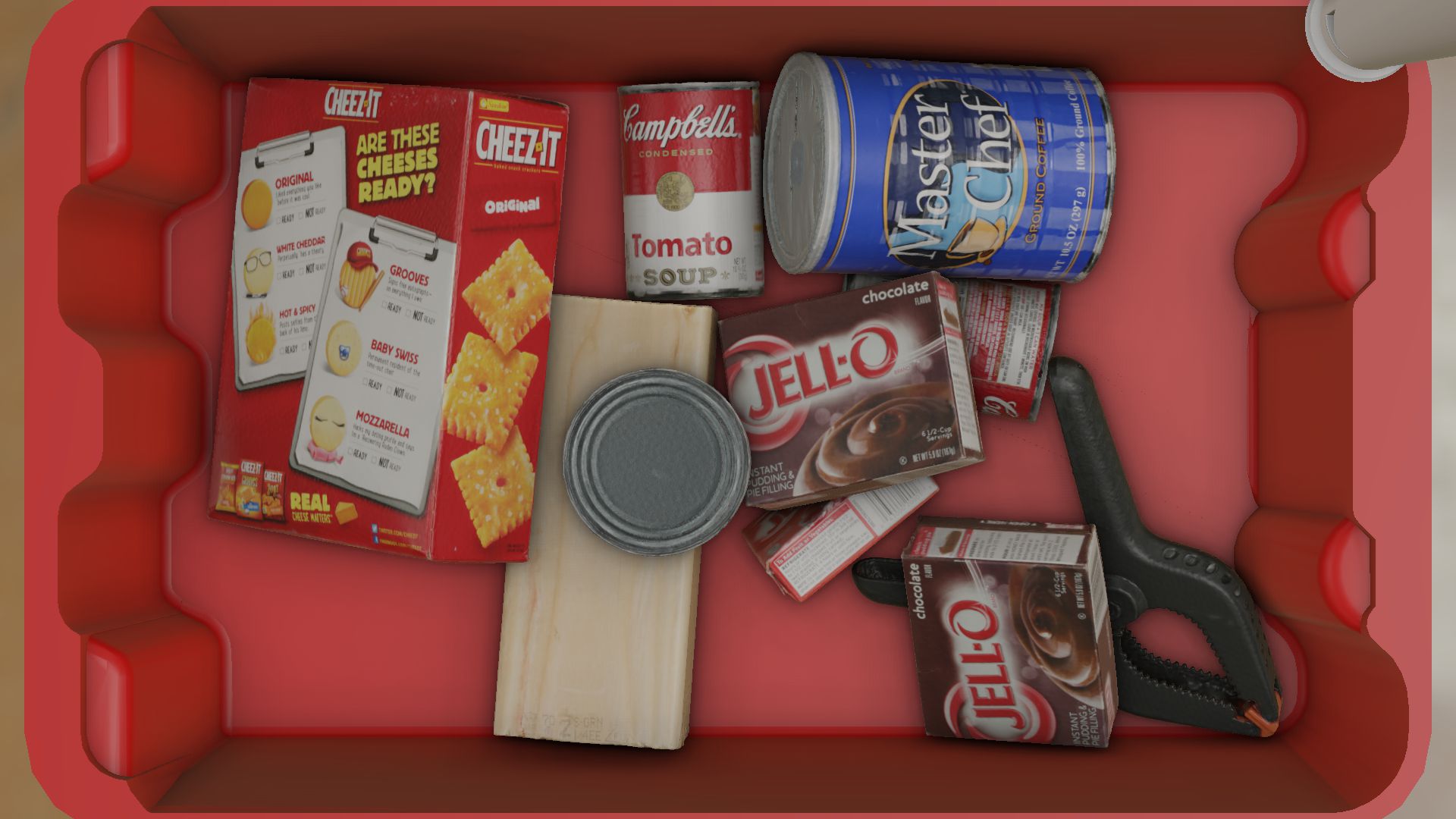} &
  \includegraphics[valign=c,width=\imgw]{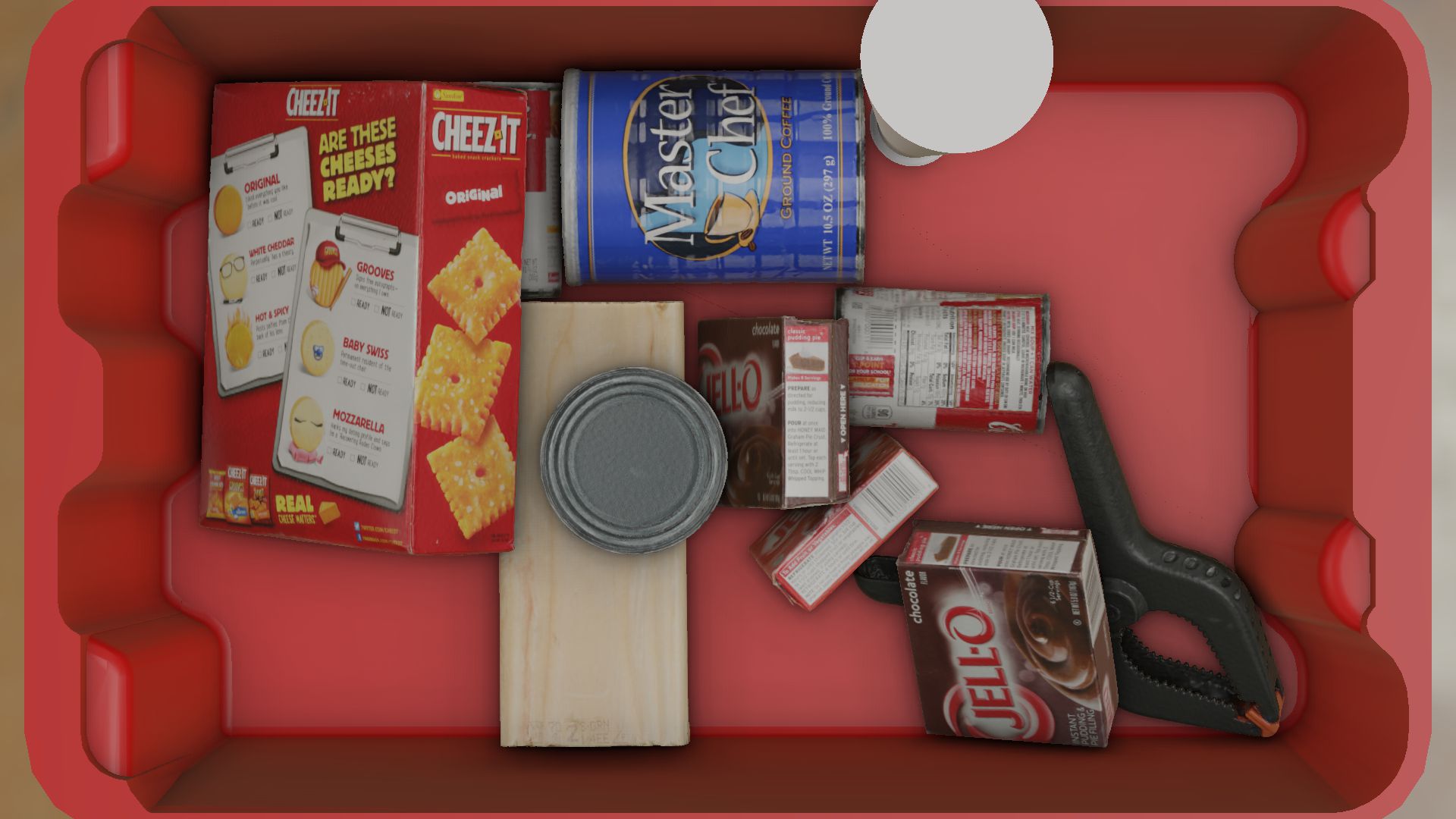} &
  \includegraphics[valign=c,width=\imgw]{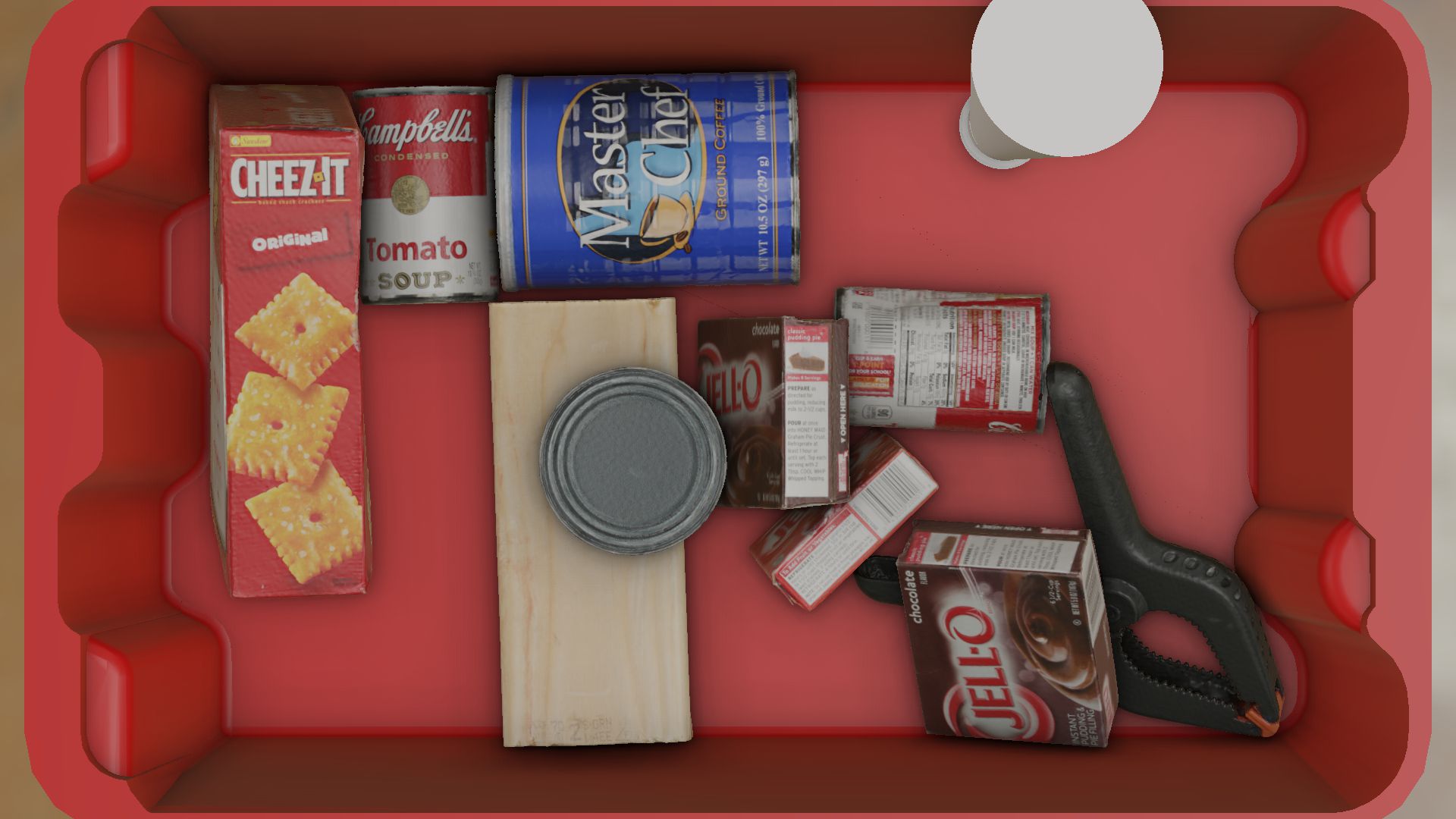} &
  \includegraphics[valign=c,width=\imgw]{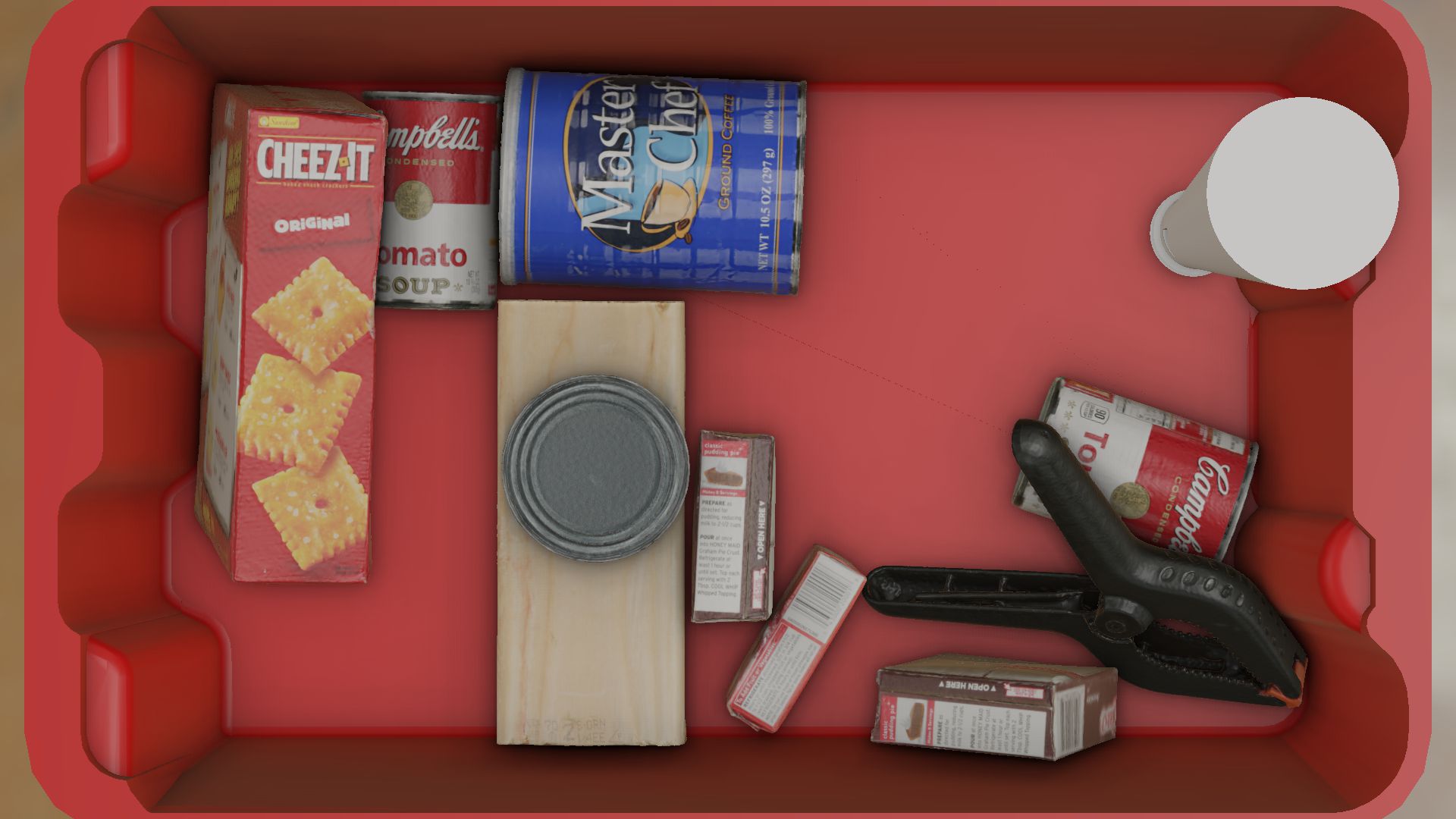} &
  \includegraphics[valign=c,width=\imgw]{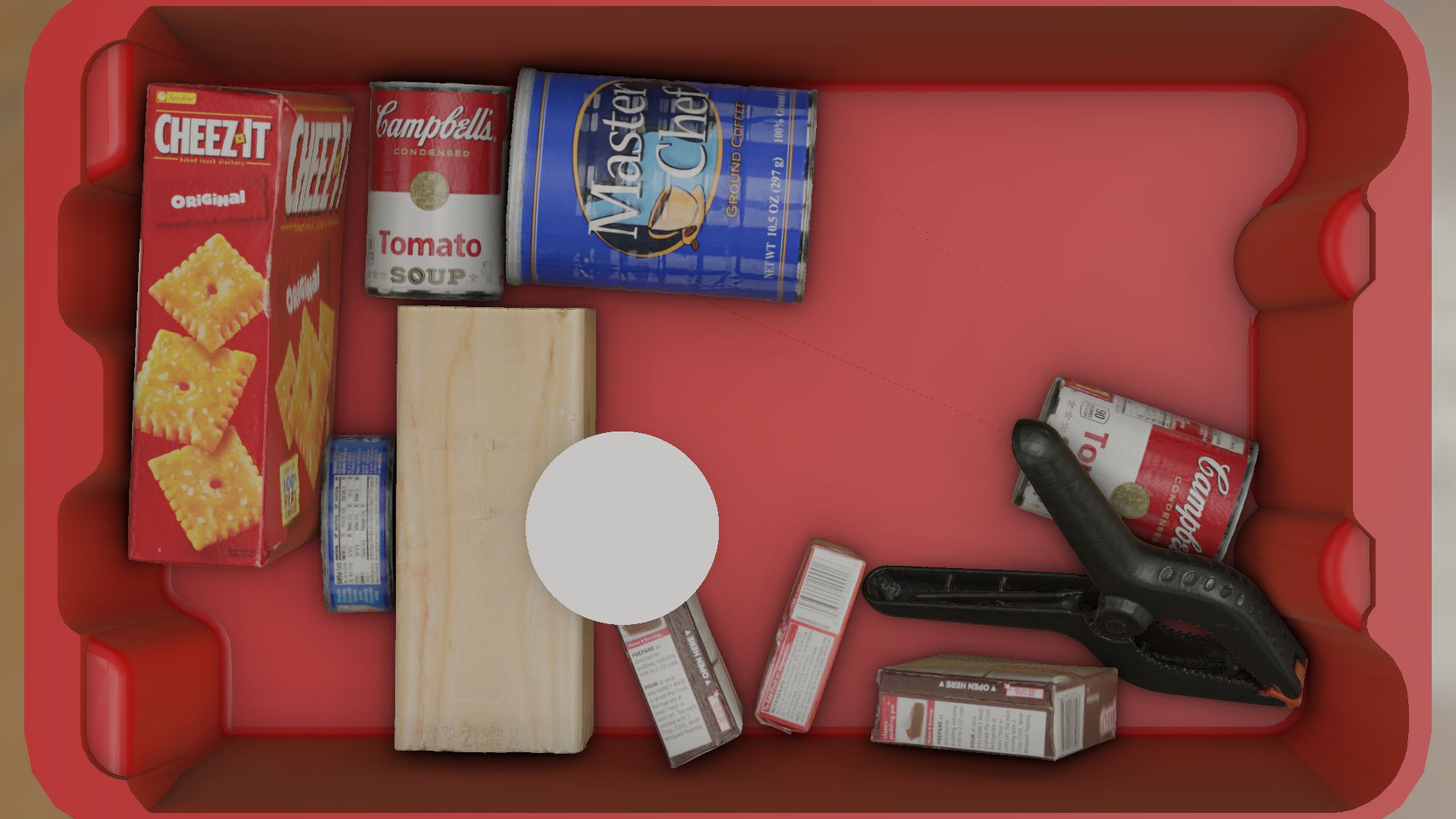} \\
  
 \end{tabular}
 \caption{Exemplary scenes from the dataset demonstrating the evolution of the scene while the gripper is performing picking (a) and moving (b) actions. Objects in the tote are covered/uncovered as the scene evolves. }
 \label{fig:examples}
\end{figure*}

Once the target position is reached, switching on the suction is simulated by performing ten raycasts in gripper
direction, emitting from the gripper perimeter. Every object hit by the raycast (with a maximum distance of 3\,cm) is
considered caught. For every caught object, a PhysX joint is created between it and the suction cup,
simulating a strong force pulling the object against the gripper as well as limiting its orientation relative to the
gripper.
The joints have a force limit of 40\,$N$ or 2\,$N$, depending on whether all rays found a target (indicating a
good vacuum seal). If the force required to keep the object at the suction cup exceeds this limit, the joint
breaks and the object is dropped back into the tote.

The picking process is repeated until the tote is empty. If three picking attempts
have failed, we also stop the sequence to prevent infinite loops.
An example sequence is shown in \cref{fig:examples}a).

\subsection{Targeted Picking}

In a second mode, we simulate \textit{targeted} picking, where a specific object
needs to be extracted. We run the same pipeline as above, with one key difference:
We choose the object which is occluded \textit{most}, i.e. is at the bottom of the
clutter graph. This choice should lead to more complex object interactions during picking.

\subsection{Moving}

As a third possible action, we perform a non-picking manipulation sequence. The goal
is to disturb the object arrangement so that other objects become visible.

We simulate the move action by moving the gripper, starting from one corner inside the tote, to
all four corners in random order. The gripper is moved at a fixed velocity of 0.1\,$\frac{m}{s}$.
In this mode, the gripper is operated with a lower stiffness of 1000\,$\frac{N}{m}$ and
a force limit of 30\,$N$. This ensures we do not squeeze objects against the immovable
tote too much, which could result in instability of the physics simulation.
An exemplary scene for the move action is shown in \cref{fig:examples}b).

\section{Dataset Statistics}

\begin{table}
  \centering
  \begin{threeparttable}
   \caption{Datasets statistics}
   \label{tab:visibility}\footnotesize
   \begin{tabular}{lcrr}
   \toprule
   Mode & Split & Frames &  Object visibility  \\
   \midrule
 Untargeted pick & train & 137,544 & 0.77 \\
 Move            & train & 99,786 & 0.67 \\
 Targeted pick   & train & 164,991 & 0.77 \\
 Untargeted pick & test  & 31,119 & 0.77 \\
 Move            & test  & 23,910 & 0.69 \\
 Targeted pick   & test & 45,882 & 0.75 \\
 \midrule
 Total & & 503232 & 0.74 \\

   \bottomrule
  \end{tabular}
  Number of frames in each SynPick dataset split along with the corresponding mean visibility fraction.
  \end{threeparttable}
\vspace{-3mm}
\end{table}

\begin {figure}
\centering
\begin{tikzpicture}
  \begin{axis}[
      ybar stacked,
      bar width=0.2cm,
      ymin=0,
      ymax=250000,
      xtick=data,
      legend columns=3,
      legend style={
          cells={anchor=west},
          at={(0.5,1.1)}, anchor=south,
          font=\scriptsize
      },
      reverse legend=true,
      xticklabels from table={\testdata}{Label},
      xticklabel style={rotate=50,anchor=east, font=\scriptsize},
      ticklabel style={font=\scriptsize},
      cycle list/Set1-6,
      height=5cm,
      width=1.1\linewidth
  ]
      \addplot+[fill]
          table [y=trainpick, meta=Label, x expr=\coordindex]
              {\testdata};
                  \addlegendentry{Untargeted pick-train}
      \addplot+[fill]
          table [y=trainmove, meta=Label, x expr=\coordindex]
              {\testdata};
                  \addlegendentry{Move-train}
      \addplot+[fill,point meta=y]
                  table [y=trainbad, meta=Label, x expr=\coordindex]
              {\testdata};
                \addlegendentry{Targeted pick-train}
      \addplot+[fill,point meta=y]
          table [y=testpick, meta=Label, x expr=\coordindex]
              {\testdata};
                  \addlegendentry{Untargeted pick-test}
      \addplot+[fill,point meta=y]
                  table [y=testmove, meta=Label, x expr=\coordindex]
              {\testdata};
                \addlegendentry{Move-test}
      \addplot+[draw=cyan!80,fill=cyan!80,point meta=y]
              table [y=testbad, meta=Label, x expr=\coordindex]
          {\testdata};
            \addlegendentry{Targeted pick-test}
  \end{axis}
\end{tikzpicture}
\label{fig:obj_dist}
\vspace{-7ex}
\caption{Number of 6D pose annotations for each object category present in SynPick dataset splits.}
\vspace{-3ex}
\end{figure}
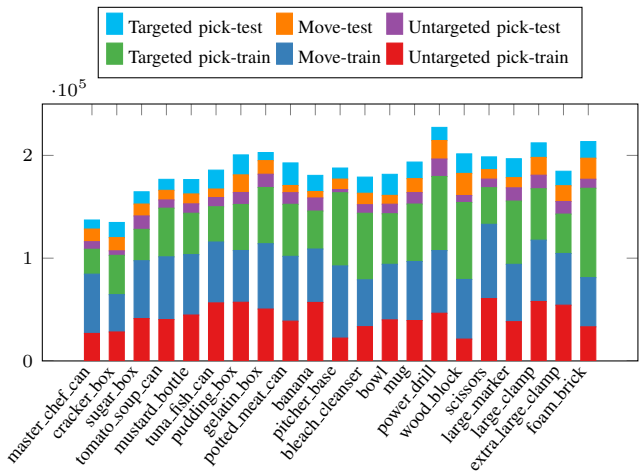

 \subsubsection{Number of Frames}
SynPick consists of 240 scenes for training and 60 scenes for testing for each of the targeted pick, untargeted pick, and move tasks.
Each scene consists of a varying number of frames. In \cref{tab:visibility}, we present the number of frames present in each split.

\subsubsection{Mean Visibility Fraction}
Like in any computer vision task, occlusions present a significant hindrance for 6D pose estimation.
A prerequisite for training robust 6D pose estimation models is a dataset that captures real-world occlusion scenarios.
Physically realistic simulation of dynamic bin picking scenes captures more realistic occlusion scenarios that are not captured
in a static scene. To analyze the degree of occlusions present in the SynPick dataset,
we present in \cref{tab:visibility}, the mean visibility fraction for the different SynPick dataset splits.
\subsubsection{Object Distribution}
As discussed in \cref{sec:scene_generaion}, the arrangement engine picks objects at random until the total volume exceeds a threshold.
In \cref{fig:obj_dist}, we present the distribution of object categories across various splits of the SynPick dataset.

\section{Baseline Object Perception Methods}

\subsection{Single-View RGB(D) 6D Pose Estimation}
\label{sec:baseline:single}

\paragraph{State-of-the-Art Methods}
The state-of-the-art methods for single-view RGB(D) pose estimation are predominantly deep-learning-based.
Some example methods include direct-regression \citep{xiang2017posecnn, wang2019densefusion, tekin2018real},
keypoint-based methods \citep{peng2019pvnet, rad2017bb8, tekin2018real, Oberweger2018}, 
render-and-compare methods \citep{li2018deepim, Shao_2020_CVPR, periyasamy2019refining}, and augmented
autoencoders \citep{Sundermeyer_2018_ECCV, deng2021poserbpf}. The BOP Challenge is organized to benchmark the progress
of 6D pose estimation methods \citep{hodavn2020bop, hodan2018bop}.

\paragraph{Baseline Method}
We evaluate CosyPose \citep{labbe2020}, the state-of-the-art winning entry of BOP Challenge 2020 \citep{hodavn2020bop}, on the SynPick dataset to establish
a baseline for our dataset. CosyPose consists of an object detection module, a coarse refinement module,
and a fine refinement module. The coarse and fine refinement modules, based on DeepIM \citep{li2018deepim}, formulate
6D pose estimation as an iterative refinement process. The coarse refinement module estimates an initial 6D pose given the 
canonical pose of the object as the input. It is trained completely on a large synthetically generated dataset. In contrast,
the fine refinement module that estimates an accurate 6D pose given the coarse 6D pose as an input is trained using the YCB-Video
dataset, disturbing the ground truth pose annotations to form inputs to be corrected by the network.
Since the coarse refinement module is trained only on a domain-agnostic synthetic dataset,
fine-tuning it on the SynPick dataset is not necessary.
Thus, we directly use the coarse refinement module weights provided by the 
authors, whereas we fine-tune the object detection and 
fine pose refinement module on the SynPick dataset.
Evaluation results on our test splits are presented in \cref{tab:experiments:results}.

\paragraph{Evaluation Metrics}
We use the area under the  accuracy-threshold curve (AUC) of ADD and ADD-S metrics for varying thresholds between 0 and 0.1m \citep{xiang2017posecnn}. 
ADD metric is the average distance between the model points in ground truth and estimated poses. Formally, 
\begin{equation}
 \text{ADD} = \frac{1}{m} \sum_{x \in \mathcal{M}} \| (\mathbf{Rx} + \mathbf{T}) - (\widetilde{\mathbf{R}}\mathbf{x}  + \widetilde{\mathbf{T}})  \|
\end{equation}
where $\mathcal{M}$ is the set of 3D model points with $m$ number of points, $\mathbf{R}$ and $\mathbf{T}$ are orientation and translation
components of ground truth 6D pose, and $\widetilde{\mathbf{R}}$ and $\widetilde{\mathbf{T}}$ are estimated orientation and translation, respectively.

Objects that exhibit symmetries perform poorly on the ADD metric. ADD-S is a variant of ADD metric that takes symmetries into account by formulating an ICP-like metric that selects closes points:
\begin{equation}
  \text{ADD-S} = \frac{1}{m} \sum_{x_1 \in \mathcal{M}} \min_{x_2 \in \mathcal{M}} \| (\mathbf{Rx_1} + \mathbf{T}) - (\widetilde{\mathbf{R}}\mathbf{x_2}  + \widetilde{\mathbf{T}})  \|.
 \end{equation}

\begin{table}
  \centering
  \caption{Evaluation metrics.}
  \label{tab:experiments:results}
  \begin{threeparttable}
  \scriptsize\centering\setlength{\tabcolsep}{.2cm}
  \begin{filecontents*}{figures/eval/results.txt}
Class CosyPoseADD CosyPoseICPADDS AlphalowADD AlphalowADDs AlphahighADD AlphahighADDs
master\_chef\_can                   0.415131  0.789910  0.488062  0.817823  0.474120  0.80364
cracker\_box                        0.815021  0.882966  0.684531  0.763840  0.647531  0.73895
sugar\_box                          0.710720  0.763949  0.727643  0.789667  0.704976  0.77214
tomato\_soup\_can                   0.423993  0.514405  0.480746  0.615210  0.443602  0.57844
mustard\_bottle                     0.698644  0.804017  0.658773  0.767943  0.647044  0.75952
tuna\_fish\_can                     0.571530  0.757738  0.626235  0.769372  0.621503  0.76546
pudding\_box                        0.602133  0.682853  0.681141  0.768717  0.650928  0.74671
gelatin\_box                        0.604473  0.646989  0.691287  0.741705  0.667832  0.72880
potted\_meat\_can                   0.700385  0.789015  0.655128  0.774993  0.636370  0.76410
banana                              0.373203  0.506681  0.383876  0.501657  0.379178  0.49674
pitcher\_base                       0.792183  0.849474  0.720432  0.802431  0.699660  0.78504
bleach\_cleanser                    0.687058  0.739984  0.704005  0.780168  0.685327  0.76898
bowl                                0.756585  0.742398  0.751215  0.751215  0.741341  0.74134
mug                                 0.646717  0.771644  0.641635  0.777484  0.617256  0.76320
power\_drill                        0.696453  0.765639  0.761633  0.832325  0.738353  0.81994
wood\_block                         0.734424  0.722977  0.746756  0.746755  0.731868  0.73186
scissors                            0.447030  0.498165  0.412011  0.469730  0.386966  0.44563
large\_marker                       0.312542  0.378977  0.368709  0.452094  0.372019  0.45449
large\_clamp                        0.567885  0.685406  0.601994  0.601994  0.591316  0.59131
extra\_large\_clamp                 0.567053  0.725005  0.594444  0.594444  0.575196  0.57519
foam\_brick                         0.546452  0.753085  0.765071  0.765071  0.747768  0.74776
\end{filecontents*}
\pgfplotstabletypeset[
    columns={Class, CosyPoseADD, CosyPoseICPADDS, AlphalowADD, AlphalowADDs, AlphahighADD, AlphahighADDs}, 
   every column/.style={
     column type=r,
     /pgf/number format/fixed,
     /pgf/number format/fixed zerofill,
     /pgf/number format/precision=1,
   },
   columns/Class/.style={
     column name={Object},
     column type=l,
     string type,
     string replace={SymC}{Class-wise average},
   },
   columns/CosyPoseADD/.style={
     column name={ADD},
     multiply with=100,
   },
   columns/CosyPoseICPADDS/.style={
     column name={ADD-S},
     multiply with=100,
   },
   columns/AlphalowADD/.style={
     column name={ADD},
     multiply with=100,
   },
   columns/AlphalowADDs/.style={
     column name={ADD-S},
     multiply with=100,
   },
   columns/AlphahighADD/.style={
     column name={ADD},
     multiply with=100,
   },
   columns/AlphahighADDs/.style={
     column name={ADD-S},
     multiply with=100,
   },
  every head row/.style={
    before row={\toprule
     & \multicolumn{2}{c}{CosyPose}
     & \multicolumn{2}{c}{Filter $\alpha$=0.05}
     & \multicolumn{2}{c}{Filter $\alpha$=0.2} \\
     \cmidrule (lr) {2-3}
     \cmidrule (lr) {4-5}
     \cmidrule (lr) {6-7}
    },
    after row={\midrule
    },
  },
  every last row/.style={after row={
  \midrule
    Mean
    & 60.3
    & 70.3
    & 62.6
    & 70.9
    & 60.8
    & 69.4 \\
  \bottomrule}},
  ]
  {figures/eval/results.txt}
  \end{threeparttable}
\vspace{-3ex}
\end{table}

\subsection{6D Pose Tracking}

\pgfplotstableread{data/filtering.txt}\filtering

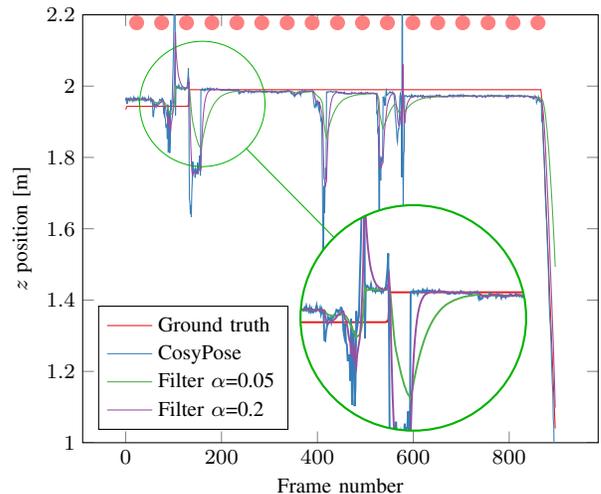
\begin{figure}[b] \centering\footnotesize
 \begin{tikzpicture}[spy using outlines={circle, magnification=1.8, size=0.5cm, connect spies},
    m/.style={below,circle,fill=red!50,inner sep=2pt}
  ]
  \begin{axis}[cycle list/Set1-4, legend pos=south west, ymax=2.2, ymin=1, legend cell align={left}, xlabel={Frame number}, ylabel={$z$ position [m]}]
   \addplot+[mark=none] table [x=t,y=gtz] {\filtering};
   \addlegendentry{Ground truth}
   
   \addplot+[mark=none] table [x=t,y=pz] {\filtering};
   \addlegendentry{CosyPose}
   
   \addplot+[mark=none] table [x=t,y=fz] {\filtering};
   \addlegendentry{Filter $\alpha$=0.05}
   
   \addplot+[mark=none] table [x=t,y=f2z] {\filtering};
   \addlegendentry{Filter $\alpha$=0.2}
   
   \coordinate (spypoint) at (axis cs:160,1.95);
   \coordinate (glass) at (axis cs:600,1.35);
   
   \node[m] at (axis cs:23, 2.2) {};
   \node[m] at (axis cs:75, 2.2) {};
   \node[m] at (axis cs:127, 2.2) {};
   \node[m] at (axis cs:180, 2.2) {};
   \node[m] at (axis cs:232, 2.2) {};
   \node[m] at (axis cs:284, 2.2) {};
   \node[m] at (axis cs:337, 2.2) {};
   \node[m] at (axis cs:389, 2.2) {};
   \node[m] at (axis cs:442, 2.2) {};
   \node[m] at (axis cs:494, 2.2) {};
   \node[m] at (axis cs:546, 2.2) {};
   \node[m] at (axis cs:599, 2.2) {};
   \node[m] at (axis cs:651, 2.2) {};
   \node[m] at (axis cs:703, 2.2) {};
   \node[m] at (axis cs:756, 2.2) {};
   \node[m] at (axis cs:808, 2.2) {};
   \node[m] at (axis cs:860, 2.2) {};
   
  \end{axis}
  
  \spy[green!70!black,size=3cm] on (spypoint) in node [fill=white] at (glass);
 \end{tikzpicture}
 \vspace{-2ex}
 \caption{Position trajectory ($z$ axis) of the \textit{scissors} object in the first test scene.
  We show raw CosyPose predictions and an exponentially moving average with different
  recursive filter coefficients $\alpha$. The time of each pick attempt has been marked with a red circle.}
 \label{fig:filtering_z}
\end{figure}

The 6D pose estimation baseline established in \cref{sec:baseline:single} is useful,
but does not really capture a real bin picking situation. In our picking scenario,
which is typical for industrial applications, a tote of objects is emptied completely,
object by object. It is certainly beneficial to monitor the object poses over time,
to make use of dependencies between frames.
This way, not only temporary effects such as occlusions by the gripper or other objects
can be mitigated, but noisy predictions can also be smoothed to obtain a more precise
estimate than from a single frame alone.

\paragraph{State of the Art}

While there are many works on single-view pose estimation, the research field
in 6D pose tracking is narrower, but just as diverse.
\citet{wang20206} present a 6D pose estimation
and tracking framework based on self-supervised sparse keypoints.
\citet{deng2021poserbpf} formulate the pose tracking problem in the particle filter framework.
In contrast, \citet{wen2020se} follow a render-and-compare approach to perform
very fast 6D pose tracking with 90\,fps.

\paragraph{Baseline}

We present a naive tracking baseline based on the CosyPose approach evaluated in
\cref{sec:baseline:single}. This baseline is intended to demonstrate the low-hanging
fruit which can be reached by temporal filtering. We implement an exponentially moving
average (EMA) with recursive filter coefficient $\alpha$:

\begin{equation}
 \hat{t}_n = (1-\alpha) \hat{t}_{n-1} + \alpha t_{n},
\end{equation}
where $t_n$ is the CosyPose translation estimate at frame $n$ of the sequence
and $\hat{t}_n$ is the filtered output.
The object orientations are filtered very similarly, but in quaternion space
in order to interpolate the orientations correctly:
\begin{equation}
 \hat{q}_n = \textrm{slerp}(\hat{q}_{n-1}, q_n, \alpha),
\end{equation}
where slerp is the spherical linear interpolation function, which interpolates
with $0<\alpha<1$ between the two given rotations.

\Cref{fig:filtering_z} shows a sequence of raw CosyPose translation predictions in the $z$ axis (into the image) for one exemplary object.
It can be seen that CosyPose exhibits both stationary noise as well as large deviations,
which are mostly caused by temporary occlusions---either by the gripper or other objects.
While our naive filtering cannot address steady-state errors, it does smoothen the
stationary noise and softens the large jumps caused by occlusions.
\Cref{tab:experiments:results} corroborates these results: Both ADD and ADD-S scores
benefit from filtering.

\section{Discussion \& Conclusion}

We presented the SynPick dynamic bin-picking dataset together with its
online generator. Our baseline experiments demonstrate that
state-of-the-art can achieve good per-frame accuracy, but also that
temporal filtering should be employed to correct both estimator noise
and effects of temporary occlusions.
Further research in the applicability of recent tracking approaches
to industrial robotics applications is definitely warranted.
We hope that our dataset can serve as an inspiration to researchers
in the field of bin picking and facilitates 6D object tracking for industrial automation.

\section*{Acknowledgment}

This research has been supported by an Amazon Research Award and by the Competence Center for Machine Learning Rhine Ruhr
(ML2R), which is funded by the Federal Ministry of Education and Research of Germany (grant no. 01—S18038A).

%
%

\printbibliography

\end{document}